\documentclass[11pt,a4paper]{article}
\usepackage[hyperref]{acl2017}
\usepackage{times}

\usepackage{hyperref}
\usepackage{url}
\let\<\textless
\let\>\textgreater

\usepackage{afterpage}
\usepackage{float}
\usepackage{lipsum}
\usepackage{latexsym}
\usepackage{amsmath}
\usepackage{amsfonts}
\usepackage{graphicx}
\usepackage{epstopdf}
\usepackage{multirow}
\usepackage{csquotes}
\usepackage{longtable}
\usepackage{enumitem}
\usepackage{rotating}
\usepackage{lscape}
\usepackage{wrapfig}
\usepackage{subcaption}
\usepackage{nicefrac}       
\usepackage{microtype}      
\usepackage{framed}
\usepackage{tabularx}
\usepackage{booktabs}

\newcommand*{\MyInd}{\hspace*{0.2cm}}%

\title{Towards an Automatic Turing Test:\\ Learning to Evaluate Dialogue Responses \vspace{3mm}}

\author{ Ryan Lowe$^{\heartsuit*}$ 
\And Michael Noseworthy$^{\heartsuit}$\thanks{\hspace{2mm}Indicates equal contribution.}
\And Iulian V. Serban$^{\diamondsuit}$ \vspace{-4mm}
\AND  Nicolas A.-Gontier$^{\heartsuit}$ 
\And Yoshua Bengio$^{\diamondsuit\ddagger}$
\And Joelle Pineau$^{\heartsuit\ddagger}$\\ 
\vspace{-6mm}
\AND 
$^\heartsuit$ \normalfont{Reasoning and Learning Lab, School of Computer Science, McGill University}\\
$^\diamondsuit$ Montreal Institute for Learning Algorithms, Universit{\'e} de  Montr{\'e}al\\
 $^\ddagger$ CIFAR Senior Fellow
 }

\aclfinalcopy

\begin{document}

\maketitle

\begin{abstract}
    Automatically evaluating the quality of dialogue responses for unstructured domains is a challenging problem. Unfortunately, existing automatic evaluation metrics are biased and correlate very poorly with human judgements of response quality. Yet having an accurate automatic evaluation procedure is crucial for dialogue research, as it allows rapid prototyping and testing of new models with fewer expensive human evaluations. In response to this challenge, we formulate automatic dialogue evaluation as a learning problem. We present an evaluation model ({\sc adem}) that learns to predict human-like scores to input responses, using a new dataset of human response scores. We show that the {\sc adem} model's predictions correlate significantly, and at a level much higher than word-overlap metrics such as BLEU, with human judgements at both the utterance and system-level. We also show that {\sc adem} can generalize to evaluating dialogue models unseen during training, an important step for automatic dialogue evaluation.\footnote{Code and pre-trained model parameters are available: \texttt{github.com/mike-n-7/ADEM}.}   

\end{abstract}


\section{Introduction}

Building systems that can naturally and meaningfully converse with humans has been a central goal of artificial intelligence since the formulation of the Turing test~\citep{turing1950computing}.
Research on one type of such systems, sometimes referred to as non-task-oriented dialogue systems, goes back to the mid-60s with Weizenbaum's famous program \textit{ELIZA}: a rule-based system mimicking a Rogerian psychotherapist by persistently either rephrasing statements or asking questions \citep{weizenbaum1966eliza}.
Recently, there has been a surge of interest towards building large-scale non-task-oriented dialogue systems using neural networks~\citep{sordoni2015neural,shang2015neural,vinyals2015neural,DBLP:conf/aaai/SerbanSBCP16,li2015diversity}.
These models are trained in an end-to-end manner to optimize a single objective, usually the likelihood of generating the responses from a fixed corpus.
Such models have already had a substantial impact in industry, including Google's Smart Reply system~\citep{kannan2016smart}, and Microsoft's Xiaoice chatbot \citep{markoff2015forsymp}, which has over 20 million users.

\begin{figure}
\fontsize{10.5}{12}\selectfont
\centering
\begin{tabular}{l}
\hline
\textbf{Context of Conversation} \\ 
Speaker A: Hey, what do you want to do tonight?\\
Speaker B: Why don't we go see a movie?\\ \hline
\textbf{Model Response}\\ 
Nah, let's do something active. \\ \hline
\textbf{Reference Response}\\
Yeah, the film about Turing looks great!\\
\hline
\end{tabular}
\caption{\label{tab:toy} Example where word-overlap scores fail for dialogue evaluation; although the model response is reasonable, it has no words in common with the reference response, and thus would be given low scores by metrics such as BLEU.}
\vspace{-5mm}
\end{figure}

One of the challenges when developing such systems is to have a good way of measuring progress, in this case the performance of the chatbot. The Turing test provides one solution to the evaluation of dialogue systems, but there are limitations with its original formulation.
The test requires live human interactions, which is expensive and difficult to scale up.
Furthermore, the test requires carefully designing the instructions to the human interlocutors, in order to balance their behaviour and expectations so that different systems may be ranked accurately by performance.
Although unavoidable, these instructions introduce bias into the evaluation measure. 
The more common approach of having humans evaluate the quality of dialogue system responses, rather than distinguish them from human responses, induces similar drawbacks in terms of time, expense, and lack of scalability. 
In the case of chatbots designed for specific conversation domains, it may also be difficult to find sufficient human evaluators with appropriate background in the topic~\citep{lowe2015ubuntu}.






Despite advances in neural network-based models,
evaluating the quality of dialogue responses automatically remains a challenging and under-studied problem in the non-task-oriented setting.
The most widely used metric for evaluating such dialogue systems is BLEU~\citep{papineni2002bleu}, a metric measuring word overlaps originally developed for machine translation.
However, it has been shown that BLEU and other word-overlap metrics are biased and correlate poorly with human judgements of response quality~\citep{liu2016not}.
There are many obvious cases where these metrics fail, as they are often incapable of considering the semantic similarity between responses (see Figure \ref{tab:toy}).
Despite this, many researchers still use BLEU to evaluate their dialogue models~\citep{ritter2011data,sordoni2015neural,li2015diversity,galley2015deltableu,li2016persona}, as there are few alternatives available that correlate with human judgements.
While human evaluation should always be used to evaluate dialogue models, it is often too expensive and time-consuming to do this for every model specification (for example, for every combination of model hyperparameters).
Therefore, having an accurate model that can evaluate dialogue response quality automatically --- what could be considered an \textit{automatic Turing test} --- is critical in the quest for building human-like dialogue agents.

To make progress towards this goal, we make the simplifying assumption that a `good' chatbot is one whose responses are scored highly on appropriateness by human evaluators. We believe this is sufficient for making progress as current dialogue systems often generate inappropriate responses. We also find empirically that asking evaluators for other metrics results in either low inter-annotator agreement, or the scores are highly correlated with appropriateness (see supp.\@ material).
Thus, we collect a dataset of appropriateness scores to various dialogue responses, and we use this dataset to train an \textit{automatic dialogue evaluation model} ({\sc adem}).
The model is trained in a semi-supervised manner using a hierarchical recurrent neural network (RNN) to predict human scores. 
We show that {\sc adem} scores correlate significantly with human judgement at both the utterance-level and system-level. 
We also show that {\sc adem} can often generalize to evaluating new models, whose responses were unseen during training, making {\sc adem} a strong first step towards effective automatic dialogue response evaluation.\footnote{Code and trained model parameters are available online: \texttt{https://github.com/mike-n-7/ADEM}.}

\section{Data Collection}
\label{sec:data}


\begin{table}
    \centering
    \small
    \begin{tabular}{l r}
         \toprule
          \# Examples & 4104 \\ 
          \# Contexts & 1026 \\\hline
         \# Training examples & 2,872 \\ 
         \# Validation examples & 616  \\ 
         \# Test examples & 616 \\ \hline
         $\kappa$ score (inter-annotator & 0.63 \\
           correlation) & \\\bottomrule       
    \end{tabular}
    \caption{Statistics of the dialogue response evaluation dataset. Each example is in the form \textit{(context, model response, reference response, human score)}.}
    \label{tab:dataset}
    \vspace{-2mm}
\end{table}
To train a model to predict human scores to dialogue responses, we first collect a dataset of human judgements (scores) of Twitter responses using the crowdsourcing platform Amazon Mechanical Turk (AMT).\footnote{All data collection was conducted in accordance with the policies of the host institutions' ethics board.}
The aim is to have accurate human scores for a variety of conversational responses --- conditioned on dialogue contexts --- which span the full range of response qualities.
For example, the responses should include both relevant and irrelevant responses, both coherent and non-coherent responses and so on.
To achieve this variety, we use candidate responses from several different models.
Following \cite{liu2016not}, we use the following 4 sources of candidate responses: (1) a response selected by a TF-IDF retrieval-based model, (2) a response selected by the Dual Encoder (DE) \citep{lowe2015ubuntu}, (3) a response generated using the hierarchical recurrent encoder-decoder (HRED) model \citep{DBLP:conf/aaai/SerbanSBCP16},
and (4) human-generated responses.
It should be noted that the human-generated candidate responses are \textit{not} the reference responses from a fixed corpus, but novel human responses that are different from the reference. 
In addition to increasing response variety, this is necessary because we want our evaluation model to learn to compare the reference responses to the candidate responses. We provide the details of our AMT experiments in the supplemental material, including additional experiments suggesting that several other metrics are currently unlikely to be useful for building evaluation models. Note that, in order to maximize the number of responses obtained with a fixed budget, we only obtain one evaluation score per dialogue response in the dataset.

To train evaluation models on human judgements, it is crucial that we obtain scores of responses that lie near the distribution produced by advanced models.
This is why we use the Twitter Corpus~\citep{ritter2011data}, as such models are pre-trained and readily available. 
Further, the set of topics discussed is quite broad --- as opposed to the very specific Ubuntu Dialogue Corpus~\citep{lowe2015ubuntu} --- and therefore the model may also be suited to other chit-chat domains.
Finally, since it does not require domain specific knowledge (e.g.\@ technical knowledge), it should be easy for AMT workers to annotate.


\section{Technical Background}

\subsection{Recurrent Neural Networks}

Recurrent neural networks (RNNs) are a type of neural network with time-delayed connections between the internal units. This leads to the formation of a \textit{hidden state} $h_t$, which is updated for every input: $h_t = f(W_{hh} h_{t-1} + W_{ih} x_t)$, where $W_{hh}$ and $W_{ih}$ are parameter matrices, $f$ is a non-linear activation function such as tanh, and $x_t$ is the input at time $t$. The hidden state allows for RNNs to better model sequential data, such as language.

In this paper, we consider RNNs augmented with long-short term memory (LSTM) units~\citep{hochreiter1997long}. LSTMs add a set of gates to the RNN that allow it to learn how much to update the hidden state. LSTMs are one of the most well-established methods for dealing with the vanishing gradient problem in recurrent networks~\citep{hochreiter1991untersuchungen,bengio1994learning}. 

\subsection{Word-Overlap Metrics}

One of the most popular approaches for automatically evaluating the quality of dialogue responses is by computing their \textit{word overlap} with the reference response.
In particular, the most popular metrics are the BLEU and METEOR scores used for machine translation, and the ROUGE score used for automatic summarization.
While these metrics tend to correlate with human judgements in their target domains, they have recently been shown to highly biased and correlate very poorly with human judgements for dialogue response evaluation~\citep{liu2016not}.
We briefly describe BLEU here, and provide a more detailed summary of word-overlap metrics in the supplemental material.

\paragraph{BLEU} BLEU~\citep{papineni2002bleu} analyzes the co-occurrences of n-grams in the reference and the proposed responses.
It computes the n-gram precision for the whole dataset, which is then multiplied by a brevity penalty to penalize short translations.
For BLEU-$N$, $N$ denotes the largest value of n-grams considered (usually $N=4$).


\paragraph{Drawbacks}
One of the major drawbacks of word-overlap metrics is their failure in capturing the semantic similarity (and other structure) between the model and reference responses when there are few or no common words.
This problem is less critical for machine translation; since the set of reasonable translations of a given sentence or document is rather small, one can reasonably infer the quality of a translated sentence by only measuring the word-overlap between it and one (or a few) reference translations. 
However, in dialogue, the set of appropriate responses given a context is much larger~\citep{artstein2009semi}; in other words, there is a very high \textit{response diversity} that is unlikely to be captured by word-overlap comparison to a single response. 

Further, word-overlap scores are computed directly between the model and reference responses.
As such, they do not consider the context of the conversation.
While this may be a reasonable assumption in machine translation,
it is not the case for dialogue;  whether a model response is an adequate substitute for the reference response is clearly context-dependent.
For example, the two responses in Figure \ref{tab:toy} are equally appropriate given the context. However, if we simply change the context to: \textit{``Have you heard of any good movies recently?''}, the model response is no longer relevant while the reference response remains valid.

\begin{figure*}
\includegraphics[width=1\linewidth]{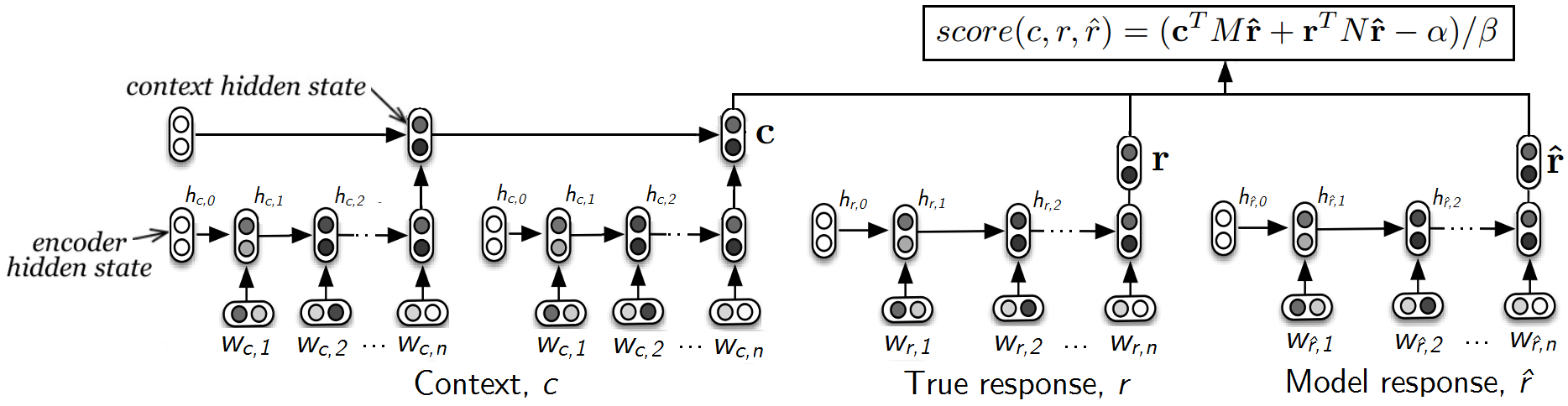}
\caption{\label{fig:model} The {\sc adem} model, which uses a hierarchical encoder to produce the context embedding $\mathbf{c}$.}
\end{figure*}

\section{An Automatic Dialogue Evaluation Model ({\sc adem})}
\label{sec:adem}

To overcome the problems of evaluation with word-overlap metrics, we aim to construct a dialogue evaluation model that: (1) captures semantic similarity beyond word overlap statistics, and (2) exploits both the context and the reference response to calculate its score for the model response.
We call this evaluation model {\sc adem}.

{\sc adem} learns distributed representations of the context, model response, and reference response using a hierarchical RNN encoder. Given the dialogue context $c$, reference response $r$, and model response $\hat{r}$, {\sc adem} first encodes each of them into vectors ($\mathbf{c}$, $\hat{\mathbf{r}}$, and $\mathbf{r}$, respectively) using the RNN encoder.
Then, {\sc adem} computes the score using a dot-product between the vector representations of $c$, $r$, and $\hat{r}$ in a linearly transformed space:
:
\begin{equation}
\label{eq:model}
score(c, r, \hat{r}) = (\mathbf{c}^T M \mathbf{\hat{r}} + \mathbf{r}^T N \mathbf{\hat{r}} - \alpha) / \beta    
\end{equation}
where $M, N \in \mathbb{R}^n$ are learned matrices initialized to the identity, and $\alpha, \beta$ are scalar constants used to initialize the model's predictions in the range $[1,5]$.
The model is shown in Figure \ref{fig:model}.

The matrices $M$ and $N$ can be interpreted as linear projections that map the model response $\mathbf{\hat{r}}$ into the space of contexts and reference responses, respectively.
The model gives high scores to responses that have similar vector representations to the context and reference response after this projection.
The model is end-to-end differentiable; all the parameters can be learned by backpropagation.
In our implementation, the parameters $\theta = \{M, N\}$ of the model are trained to minimize the squared error between the model predictions and the human score, with L2-regularization:
\begin{equation}
 \mathcal{L} = \sum_{i=1:K} [score(c_i, r_i, \hat{r}_i) - human_i]^2 + \gamma ||\theta||_2
\end{equation}
where $\gamma$ is a scalar constant. The simplicity of our model leads to both accurate predictions and fast evaluation (see supp.\@ material),  which is important to allow rapid prototyping of dialogue systems.

The hierarchical RNN encoder in our model consists of two layers of RNNs~\citep{el1995hierarchical,sordoni2015hierarchical}. 
The lower-level RNN, the \textit{utterance-level encoder}, takes as input words from the dialogue, and produces a vector output at the end of each utterance. The \textit{context-level encoder} takes the representation of each utterance as input and outputs a vector representation of the context. 
This hierarchical structure is useful for incorporating information from early utterances in the context~\citep{DBLP:conf/aaai/SerbanSBCP16}. Following previous work, we take the last hidden state of the context-level encoder as the vector representation of the input utterance or context.  The parameters of the RNN encoder are pretrained and are not learned from the human scores.

An important point is that the {\sc adem} procedure above \textit{is not a dialogue retrieval model}: the fundamental difference 
is that {\sc adem} has access to the reference response. Thus, {\sc adem} can compare a model's response to a known good response, which is significantly easier than inferring response quality from solely the context.

\paragraph{Pre-training with VHRED }
We would like an evaluation model that can make accurate predictions from few labeled examples, since these examples are expensive to obtain.
We therefore employ semi-supervised learning, and use a pre-training procedure to learn the parameters of the encoder.
In particular, we train the encoder as part of a neural dialogue model; we attach a third \textit{decoder RNN} that takes the output of the encoder as input, and train it to predict the next utterance of a dialogue conditioned on the context.

\begin{figure*}[h]
\centering
\includegraphics[width=0.8\linewidth]{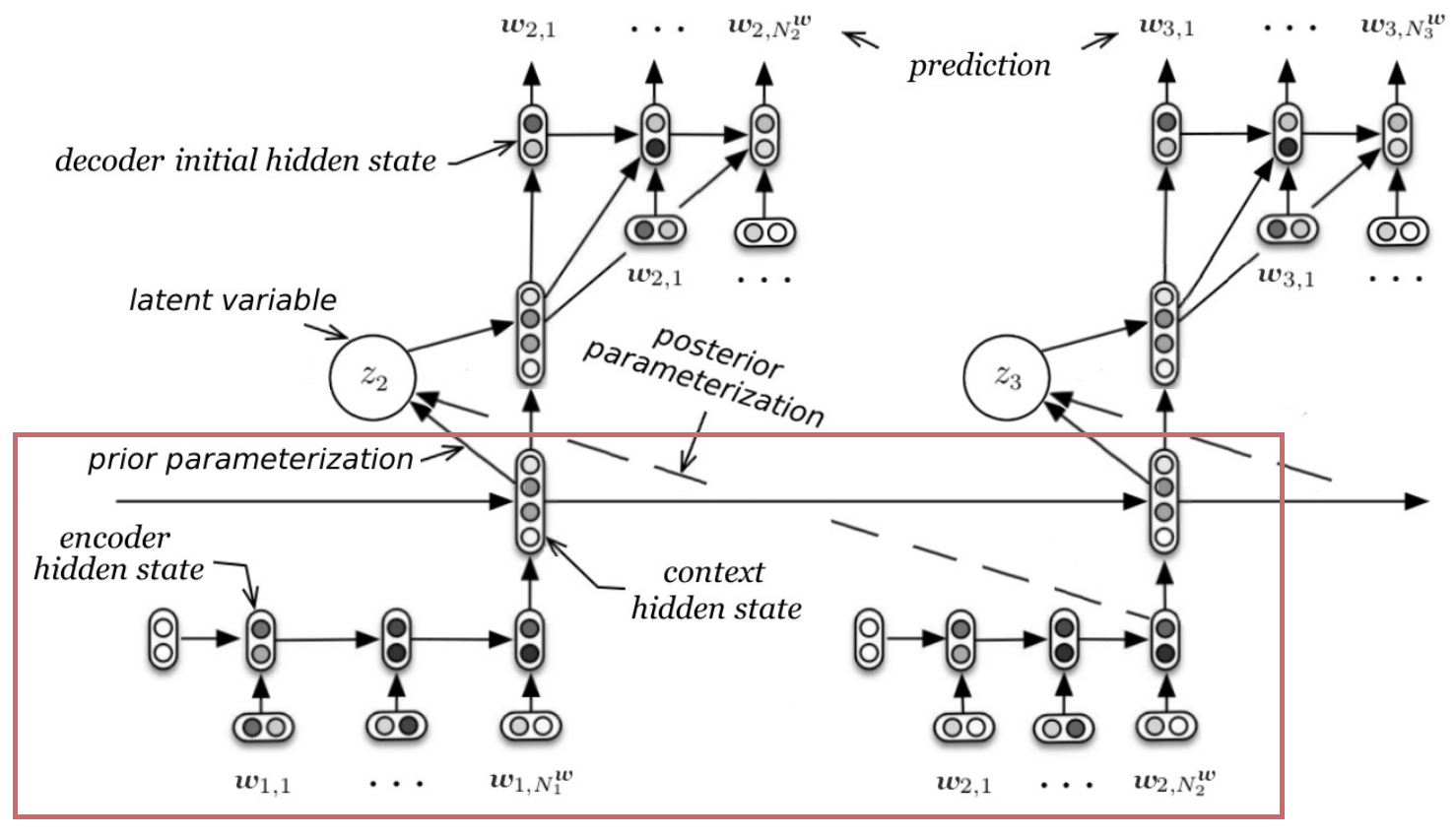}
\caption{\label{fig:vhred}The VHRED model used for pre-training. The hierarchical structure of the RNN encoder is shown in the red box around the bottom half of the figure. After training using the VHRED procedure, the last hidden state of the context-level encoder is used as a vector representation of the input text.} 
\end{figure*}

The dialogue model we employ for pre-training is the latent variable hierarchical recurrent encoder-decoder (VHRED) model~\citep{serban2016hierarchical}, shown in Figure \ref{fig:vhred}.
The VHRED model is an extension of the original hierarchical recurrent encoder-decoder (HRED) model~\citep{DBLP:conf/aaai/SerbanSBCP16} with a turn-level stochastic latent variable. 
The dialogue context is encoded into a vector using our hierarchical encoder, and the VHRED then samples a Gaussian variable that is used to condition the decoder (see supplemental material for further details).
After training VHRED, we use the last hidden state of the context-level encoder, when $c$, $r$, and $\hat{r}$ are fed as input, as the vector representations for $\mathbf{c}, \mathbf{r}$, and $\mathbf{\hat{r}}$, respectively.
We use representations from the VHRED model as it produces more diverse and coherent responses compared to HRED.

\section{Experiments}
\label{sec:experiments}

\afterpage{%
\begin{table*}[h!]
\small
\centering
\begin{tabular}{l c c c c}
\toprule
 & \multicolumn{2}{c}{\textbf{Full dataset}} & \multicolumn{2}{c}{\textbf{Test set}} \\ \hline
Metric & \textbf{Spearman }& \textbf{Pearson} & \textbf{Spearman }& \textbf{Pearson}\\ \hline

BLEU-2 & 0.039~ (0.013) & 0.081~ (\<0.001) & 0.051~ (0.254) & 0.120~ (\<0.001)    \\
BLEU-4  &  0.051~ (0.001) & 0.025~ (0.113)  & 0.063~ (0.156) & 0.073~ (0.103)    \\
ROUGE  &  0.062~ (\<0.001) & 0.114~ (\<0.001)  & 0.096~ (0.031) & 0.147~ (\<0.001)  \\ 
METEOR  &  0.021~ (0.189) & 0.022~ (0.165)  &  0.013~ (0.745) & 0.021~ (0.601)  \\
T2V & 0.140~ (\<0.001) & 0.141~ (\<0.001)  & 0.140~ (\<0.001) & 0.141~ (\<0.001)\\
VHRED &  -0.035~ (0.062) & -0.030~ (0.106)  & -0.091~ (0.023) & -0.010~ (0.805) \\ \hline
& \multicolumn{2}{c}{\textbf{Validation set}} & \multicolumn{2}{c}{\textbf{Test set}} \\ \hline
C-{\sc adem} & 0.338~ (\<0.001) & 0.355~ (\<0.001)  & 0.366~ (\<0.001) & 0.363~ (\<0.001) \\ 
R-{\sc adem} & 0.404~ (\<0.001) & 0.404~ (\<0.001)  & 0.352~ (\<0.001) & 0.360~ (\<0.001) \\
{\sc adem} (T2V) & 0.252~ (\<0.001) & 0.265~ (\<0.001) & 0.280~ (\<0.001) & 0.287 (\<0.001)  \\
{\sc adem} & \textbf{0.410}~ (\<0.001) & \textbf{0.418}~ (\<0.001) & \textbf{0.428}~ (\<0.001) & \textbf{0.436}~ (\<0.001) \\

\bottomrule
\end{tabular}
\caption{\label{tab:correlation}  Correlation between metrics and human judgements, with p-values shown in brackets. 
`{\sc adem} (T2V)' indicates {\sc adem} with tweet2vec embeddings~\citep{dhingra2016tweet2vec}, and `VHRED' indicates the dot product of VHRED embeddings (i.e.\@ {\sc adem} at initialization). C- and R-{\sc adem} represent the {\sc adem} model trained to only compare the model response to the context or reference response, respectively. We compute the baseline metric scores (top) on the full dataset to provide a more accurate estimate of their scores (as they are not trained on a training set).}
\vspace{-3mm}
\end{table*}
}

\begin{figure*}[h!]
\centering
\begin{subfigure}[b]{0.32\textwidth}
            \centering
            \includegraphics[width=.95\linewidth]{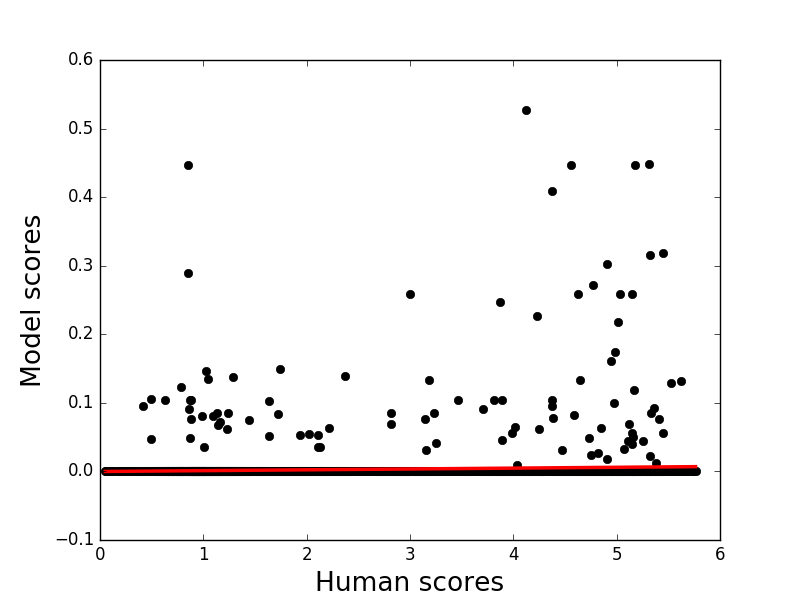}
            \caption{BLEU-2}
\end{subfigure}
\begin{subfigure}[b]{0.32\textwidth}
            \centering
            \includegraphics[width=.95\linewidth]{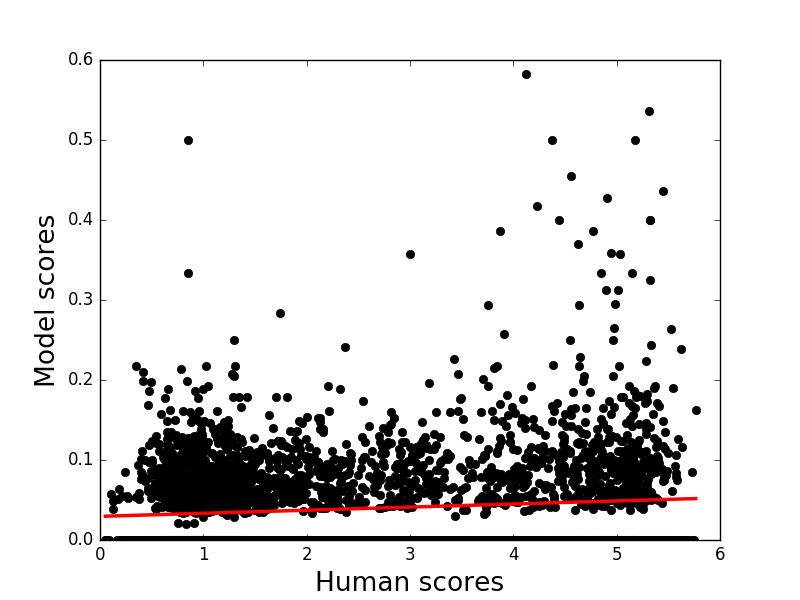}
            \caption{ROUGE}
\end{subfigure}
\begin{subfigure}[b]{0.32\textwidth}
            \centering
            \includegraphics[width=.95\linewidth]{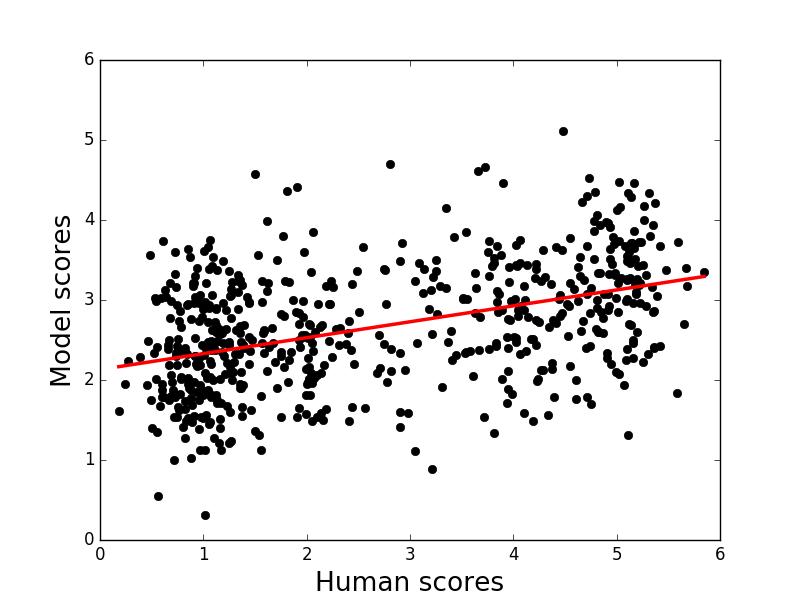}
            \caption{{\sc adem}}
\end{subfigure}\\
\caption{\label{fig:scatter} Scatter plot showing model against human scores, for BLEU-2 and ROUGE on the full dataset, 
and {\sc adem} on the test set. We add Gaussian noise drawn from $\mathcal{N}(0, 0.3)$ to the integer human scores to better visualize the density of points, at the expense of appearing less correlated.  }
\vspace{-1mm}
\end{figure*}

\subsection{Experimental Procedure} 


In order to reduce the effective vocabulary size, we use byte pair encoding (BPE) \citep{gage1994new,sennrich2015neural}, which splits each word into sub-words or characters. 
We also use layer normalization~\citep{ba2016layer} for the hierarchical encoder, which we found worked better at the task of dialogue generation than the related recurrent batch normalization~\citep{ioffe2015batch,cooijmans2016recurrent}. 
To train the VHRED model, we employed several of the same techniques found in \cite{serban2016hierarchical} and \cite{bowman2015generating}: we drop words in the decoder with a fixed rate of 25\%, and we anneal the KL-divergence term linearly from 0 to 1 over the first 60,000 batches. We use Adam as our optimizer~\citep{kingma2014adam}.

When training {\sc adem}, we also employ a sub-sampling procedure based on the model response length. In particular, we divide the training examples into bins based on the number of words in a response and the score of that response. We then over-sample from bins across the same score to ensure that {\sc adem} does not use response length to predict the score. This is because humans have a tendency to give a higher rating to shorter responses than to longer responses \citep{serban2016hierarchical}, as shorter responses are often more generic and thus are more likely to be suitable to the context. Indeed, the test set Pearson correlation between response length and human score is 0.27.


For training VHRED, we use a context embedding size of 2000. However, we found the {\sc adem} model learned more effectively when this embedding size was reduced. Thus, after training VHRED, we use principal component analysis (PCA)~\citep{pearson1901principal} to reduce the dimensionality of the context, model response, and reference response embeddings to $n$.
We found experimentally that $n=50$ provided the best performance.

When training our models, we conduct early stopping on a separate validation set. 
For the evaluation dataset, we split the train/ validation/ test sets such that there is no context overlap (i.e.\@ the contexts in the test set are unseen during training). 



\subsection{Results}

\paragraph{Utterance-level correlations }

We first present new utterance-level correlation results\footnote{We present both the Spearman correlation (computed on ranks, depicts monotonic relationships) and Pearson correlation (computed on true values, depicts linear relationships) scores.} for existing word-overlap metrics, in addition to results with embedding baselines and {\sc adem}, in Table \ref{tab:correlation}. The baseline metrics are evaluated on the entire dataset of 4,104 responses to provide the most accurate estimate of the score.  
\footnote{Note that our word-overlap correlation results in Table \ref{tab:correlation} are also lower than those presented in \cite{galley2015deltableu}. This is because Galley et al. measure corpus-level correlation, i.e.\@ correlation averaged across different subsets (of size 100) of the data, and pre-filter for high-quality reference responses.} We measure the correlation for {\sc adem} on the validation and test sets, which constitute 616 responses each. 

We also conduct an analysis of the response data from \cite{liu2016not}, where the pre-processing is standardized by removing `\<first\_speaker\>' tokens at the beginning of each utterance. The results are detailed in the supplemental material.
We can observe from both this data, and the new data in Table \ref{tab:correlation}, that the correlations for the word-overlap metrics are even lower than estimated in previous studies~\citep{liu2016not,galley2015deltableu}.
In particular, this is the case for BLEU-4, which has frequently been used for dialogue response evaluation~\citep{ritter2011data,sordoni2015neural,li2015diversity,galley2015deltableu,li2016persona}.

\begin{figure*}
    \centering
    \includegraphics[scale=0.225]{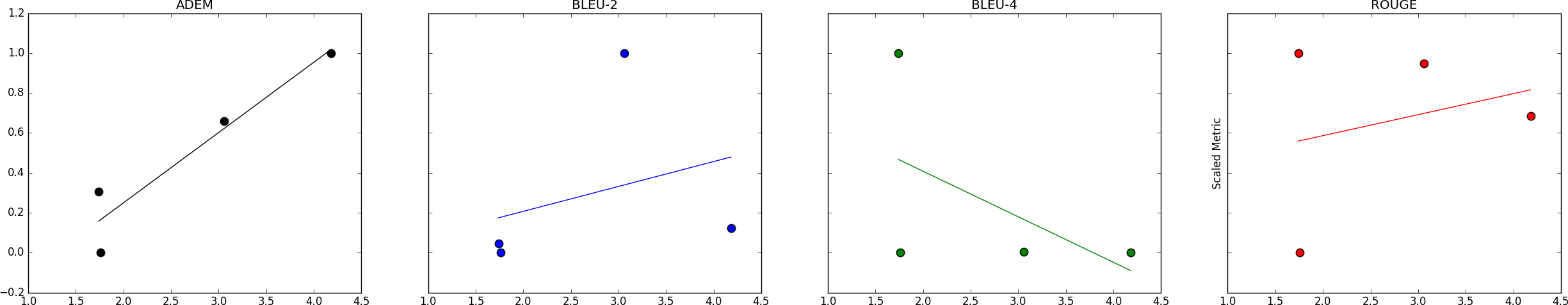}
    \caption{\label{fig:systemlevel}Scatterplots depicting the system-level correlation results for {\sc adem}, BLEU-2, BLEU-4,and ROUGE  on the test set. Each point represents the average scores for the responses from a dialogue model (TFIDF, DE, HRED, human). Human scores are shown on the horizontal axis, with normalized metric scores on the vertical axis. The ideal metric has a perfectly linear relationship.}  
\end{figure*}

We can see from Table \ref{tab:correlation} that {\sc adem} correlates far better with human judgement than the word-overlap baselines.
This is further illustrated by the scatterplots in Figure \ref{fig:scatter}. 
We also compare with {\sc adem} using tweet2vec embeddings~\citep{dhingra2016tweet2vec}. In this case, instead of using the VHRED pre-training method presented in Section \ref{sec:adem}, we use off-the-shelf embeddings for $\mathbf{c}, \mathbf{r}$, and $\mathbf{\hat{r}}$, and fine-tune $M$ and $N$ on our dataset. These tweet2vec embeddings are computed at the character-level with a bidirectional GRU on a Twitter dataset for hashtag prediction~\citep{dhingra2016tweet2vec}. We find that they obtain reasonable but inferior performance compared to using VHRED embeddings.



\paragraph{System-level correlations } 

We show the system-level correlations for various metrics in Table \ref{tab:systemlevel}, and present it visually in Figure \ref{fig:systemlevel}.
Each point in the scatterplots represents a dialogue model; humans give low scores to TFIDF and DE responses, higher scores to HRED and the highest scores to other human responses. 
It is clear that existing word-overlap metrics are incapable of capturing this relationship for even 4 models.
This renders them completely deficient for dialogue evaluation.
However, {\sc adem} produces almost the same model ranking as humans, achieving a significant Pearson correlation of 0.954.\footnote{For comparison, BLEU achieves a system-level correlation of 0.99 on 5 models in the translation domain~\citep{papineni2002bleu}.}
Thus, {\sc adem} correlates well with humans both at the response and system level.

\paragraph{Generalization to previously unseen models} When {\sc adem} is used in practice, it will take as input responses from a new model that it has not seen during training. Thus, it is crucial that {\sc adem} correlates with human judgements for new models.
We test {\sc adem}'s generalization ability by performing a leave-one-out evaluation.
For each dialogue model that was the source of response data for training {\sc adem} (TF-IDF, Dual Encoder, HRED, humans), we conduct an experiment where we train on all model responses \textit{except} those from the chosen model, and test \textit{only} on the model that was unseen during training.

The results are given in Table \ref{tab:generalization}.
We observe that the {\sc adem} model is able to generalize for all models except the Dual Encoder. This is particularly surprising for the HRED model; in this case, {\sc adem} was trained only on responses that were written by humans (from retrieval models or human-generated), but is able to generalize to responses produced by a generative neural network model. When testing on the entire test set, the model achieves comparable correlations to the {\sc adem} model that was trained on 25\% less data selected at random.  

\paragraph{Qualitative Analysis }  
To illustrate some strengths and weaknesses of {\sc adem}, we show human and {\sc adem} scores for each of the responses to various contexts in Table \ref{tab:qualitative}.
There are several instances where {\sc adem} predicts accurately: 
in particular, {\sc adem} is often very good at assigning low scores to poor responses. This seen in the first two contexts, where most of the responses given a score of 1 from humans are given scores less than 2 by {\sc adem}.
The single exception in response (4) for the second context seems somewhat appropriate and should perhaps have been scored higher by the human evaluator.
There are also several instances where the model assigns high scores to suitable responses, as in the first two contexts.

\begin{table}
\vspace{-4mm}
     \small
    \centering
        \begin{tabular}{l c}
            \toprule    
            \textbf{Metric} & \textbf{Pearson} \\ \hline
            BLEU-1  &  -0.079~ (0.921) \\
            BLEU-2 & 0.308~ (0.692)   \\
            BLEU-3 & -0.537~ (0.463)    \\
            BLEU-4 & -0.536~ (0.464)    \\
            ROUGE &  0.268~ (0.732)   \\ \hline 
            {\sc adem} & \textbf{0.954}~ (0.046) \\            
           \bottomrule         
\end{tabular}
\vspace{-2mm}
\caption{\label{tab:systemlevel}System-level correlation, with the p-value in brackets.}
\vspace{-4mm}
\end{table}

\begin{table*}[ht!]
\small
\centering
\begin{tabular}{l c c c c }
\toprule
 & \multicolumn{2}{c}{\textbf{Test on full dataset}} & \multicolumn{2}{c}{\textbf{Test on removed model responses}} \\ \hline 
Data Removed & \textbf{Spearman }&  \textbf{Pearson} &  \textbf{Spearman }&  \textbf{Pearson}  \\ \hline
 TF-IDF   &     0.406~ (\<0.001)  & 0.409~ (\<0.001) &    0.186~ (0.021)  & 0.196~ (0.015) \\
 Dual Encoder  &    0.364~ (\<0.001)  & 0.373~ (\<0.001)  & 0.026~ (0.749) & 0.027~ (0.736) \\
 HRED  &    0.393~ (\<0.001)  & 0.396~ (\<0.001)  &  0.151~ (0.060) & 0.148~ (\<0.070) \\
 Human  &    0.292~ (\<0.001) & 0.298~ (\<0.001)  &  0.216~ (\<0.010) & 0.148~ (\<0.070) \\ 
 \textbf{Average} & \textbf{0.364} & \textbf{0.369} & \textbf{0.145} & \textbf{0.130} \\ \hline
 25\% at random  &    0.378~ (\<0.001) & 0.384~ (\<0.001)  &  --- & --- \\ 
\bottomrule
\end{tabular}
\caption{\label{tab:generalization}  Correlation for {\sc adem} when various model responses are removed from the training set. The left two columns show performance on the entire test set, and the right two columns show performance on responses only from the dialogue model not seen during training. The last row (25\% at random) corresponds to the {\sc adem} model trained on all model responses, but with the same amount of training data as the model above (i.e.\@ 25\% less data than the full training set).}
\end{table*}

\begin{table*}
\scriptsize
\centering
\begin{tabular}{p{40mm} p{23mm} p{65mm} p{6mm} p{6mm}}
\toprule
\textbf{Context} & \textbf{Reference response} & \textbf{Model responses} & \textbf{Human score} & \textbf{{\sc adem} score} \\ \hline
photo to see my television debut go to - some. some on \<url\> - hehe \<url\> $\rightarrow$
{ \color{blue} it really was you? i thought ppl were recognizing someone who looked like you! were the oysters worth the wait?} 
& yeah it was me . haha i 'd kinda forgotten about it it was filmed a while ago
&  1) i'm not sure. i just don't know what to do with it. \newline 2) you heard the horsepower productions remix of lee scratch \MyInd perry's `exercising' off his `mighty upsetter' album? \newline 3) you wont chug a fuzzy peach navel \newline 4) they were! & 3 \newline 1 \newline\newline 1 \newline 5&
3.413 \newline 1.644 \newline\newline 1.944 \newline 3.085 
\\ \hline

just beat call of duty!! $\rightarrow$ {\color{blue} want a cookie?} $\rightarrow$ yes!! $\rightarrow$ {\color{blue}come get it} & im in kenmore at the moment & 1) i'm gonna get a new phone some moro \newline  2) no way man. \newline 3) wow i just got a free pizza coupon! get yours \newline \MyInd  before theres no more! \<url\> \newline 4) i'm going to go to the mall. & 1 \newline 5 \newline 1 \newline \newline 1 & 2.947 \newline 3.405 \newline 2.877 \newline\newline 2.851 \\ \hline


 am i out of twitter jail yet? testing $\rightarrow$ {\color{blue}yeah. i posted bail} $\rightarrow$ thanks. i am a right chatter tweetbox on sundays. same happened last sunday lol & 
 any news on meeting our \@user ? i go to the us on friday and i don 't want to miss anything arranged &
 1) i'm not sure if i'm going to be able to get it. \newline  2) good to see another mac user in the leadership ranks \newline 3) awww poor baby hope u get to feeling better soon. \MyInd maybe some many work days at piedmont \newline 4) did you tweet too much? &
 3 \newline 4 \newline\newline 2 \newline 5 & 
 2.651 \newline 2.775 \newline\newline 2.173 \newline 3.185 \\

\bottomrule
\end{tabular}
\vspace{-1mm}
\caption{\label{tab:qualitative}  Examples of scores given by the {\sc adem} model. }
\end{table*}

One drawback we observed is that {\sc adem} tends to be too conservative when predicting response scores. This is the case in the third context, where the model assigns low scores to most of the responses that a human rated highly. This behaviour is likely due to the squared error loss used to train {\sc adem}; since the model receives a large penalty for incorrectly predicting an extreme value, it learns to predict scores closer to the average human score. 
We provide many more experiments, including investigation of evaluation speed, learning curves, data efficiency, a failure analysis, and the primary source of improvement over word-overlap metrics in the supplemental material.


\section{Related Work}


Related to our approach is the literature on novel methods for the evaluation of machine translation systems, especially through the WMT evaluation task~\citep{callison2011findings,machacek2014results,stanojevic2015results}. In particular, \cite{albrecht2007regression,gupta2015reval} have proposed to evaluate machine translation systems using Regression and Tree-LSTMs respectively. Their approach differs from ours as, in the dialogue domain, we must additionally condition our score on the context of the conversation, which is not necessary in translation. 

There has also been related work on estimating the quality of responses in chat-oriented dialogue systems. \cite{devault2011toward} train an automatic dialogue policy evaluation metric from 19 structured role-playing sessions, enriched with paraphrases and external referee annotations. \cite{gandhe2016semi} propose a semi-automatic evaluation metric for dialogue coherence, similar to BLEU and ROUGE, based on `wizard of Oz' type data.\footnote{In `wizard of Oz' scenarios, humans play the role of the dialogue system, usually unbeknown to the interlocutors.} \cite{xiang2014problematic} propose a framework to predict utterance-level problematic situations in a dataset of Chinese dialogues using intent and sentiment factors. Finally, \cite{higashinaka2014evaluating} train a classifier to distinguish user utterances from system-generated utterances using various dialogue features, such as dialogue acts, question types, and predicate-argument structures. 

Several recent approaches use hand-crafted reward features to train dialogue models using reinforcement learning (RL). For example, \cite{li2016deep} use features related to ease of answering and information flow, and \cite{yu2016strategy} use metrics related to turn-level appropriateness and conversational depth. 
These metrics are based on hand-crafted features, which only capture a small set of relevant aspects; this inevitably leads to sub-optimal performance, and it is unclear whether such objectives are preferable over retrieval-based cross-entropy or word-level maximum log-likelihood objectives.
Furthermore, many of these metrics are computed at the conversation-level, and are not available for evaluating single dialogue responses. The metrics that can be computed at the response-level could be incorporated into our framework, for example by adding a term to equation \ref{eq:model} consisting of a dot product between these features and a vector of learned parameters.

There has been significant work on evaluation methods for task-oriented dialogue systems, which attempt to solve a user's task such as finding a restaurant. These methods include the PARADISE framework~\citep{walker1997paradise} and MeMo~\citep{moller2006memo}, which consider a task completion signal. PARADISE in particular is perhaps the first work on learning an automatic evaluation function for dialogue, accomplished through linear regression. However, PARADISE requires that one can measure task completion and task complexity, which are not available in our setting.

\section{Discussion}


We use the Twitter Corpus to train our models as it contains a broad range of non-task-oriented conversations and it has been used to train many state-of-the-art models.
However, our model could easily be extended to other general-purpose datasets, such as Reddit, once similar pre-trained models become publicly available.
Such models are necessary even for creating a test set in a new domain, which will help us determine if {\sc adem} generalizes to related dialogue domains. We leave investigating the domain transfer ability of {\sc adem} for future work.


The evaluation model proposed in this paper favours dialogue models that generate responses that are rated as highly appropriate by humans. 
It is likely that this property does not fully capture the desired end-goal of chatbot systems. 
For example, one issue with building models to approximate human judgements of response quality is the problem of generic responses. Since humans often provide high scores to generic responses due to their appropriateness for many given contexts~\citep{shang2016overview}, a model trained to predict these scores will exhibit the same behaviour. An important direction for future work is modifying {\sc adem} such that it is not subject to this bias. This could be done, for example, by censoring {\sc adem}'s representations \citep{edwards2015censoring} such that they do not contain any information about length. 
Alternatively, one can combine this with an \textit{adversarial evaluation model} \citep{kannan2017adversarial,li2017learning} that assigns a score based on how easy it is to distinguish the dialogue model responses from human responses.
In this case, a model that generates generic responses will easily be distinguishable and obtain a low score.

An important direction of future research is building models that can evaluate the capability of a dialogue system to have an engaging and meaningful interaction with a human.
Compared to evaluating a single response, this evaluation is arguably closer to the end-goal of chatbots.
However, such an evaluation is extremely challenging to do in a completely automatic way.
We view the evaluation procedure presented in this paper as an important step towards this goal; current dialogue systems are incapable of generating responses that are rated as highly appropriate by humans, and we believe our evaluation model will be useful for measuring and facilitating progress in this direction.


\subsection*{Acknowledgements}

We'd like to thank Casper Liu for his help with the correlation code, Laurent Charlin for helpful discussions on the data collection, Jason Weston for suggesting improvements in the experiments, and Jean Harb and Emmanuel Bengio for their debugging aid. We gratefully acknowledge support from the Samsung Institute of Advanced Technology, the National Science and Engineering Research Council, and Calcul Quebec. We'd also like to thank the developers of Theano~\citep{team2016theano}.


\bibliography{acl_2017}

\begin{thebibliography}{}
\expandafter\ifx\csname natexlab\endcsname\relax\def\natexlab#1{#1}\fi

\bibitem[{Albrecht and Hwa(2007)}]{albrecht2007regression}
Joshua Albrecht and Rebecca Hwa. 2007.
\newblock Regression for sentence-level mt evaluation with pseudo references.
\newblock In {\em ACL\/}.

\bibitem[{Artstein et~al.(2009)Artstein, Gandhe, Gerten, Leuski, and
  Traum}]{artstein2009semi}
Ron Artstein, Sudeep Gandhe, Jillian Gerten, Anton Leuski, and David Traum.
  2009.
\newblock Semi-formal evaluation of conversational characters.
\newblock In {\em Languages: From Formal to Natural\/}, Springer, pages 22--35.

\bibitem[{Ba et~al.(2016)Ba, Kiros, and Hinton}]{ba2016layer}
Jimmy~Lei Ba, Jamie~Ryan Kiros, and Geoffrey~E Hinton. 2016.
\newblock Layer normalization.
\newblock {\em arXiv preprint arXiv:1607.06450\/} .

\bibitem[{Banerjee and Lavie(2005)}]{banerjee2005meteor}
Satanjeev Banerjee and Alon Lavie. 2005.
\newblock Meteor: An automatic metric for mt evaluation with improved
  correlation with human judgments.
\newblock In {\em Proceedings of the acl workshop on intrinsic and extrinsic
  evaluation measures for machine translation and/or summarization\/}.
  volume~29, pages 65--72.

\bibitem[{Bengio et~al.(1994)Bengio, Simard, and Frasconi}]{bengio1994learning}
Yoshua Bengio, Patrice Simard, and Paolo Frasconi. 1994.
\newblock Learning long-term dependencies with gradient descent is difficult.
\newblock {\em IEEE transactions on neural networks\/} 5(2):157--166.

\bibitem[{Bowman et~al.(2016)Bowman, Vilnis, Vinyals, Dai, Jozefowicz, and
  Bengio}]{bowman2015generating}
Samuel~R Bowman, Luke Vilnis, Oriol Vinyals, Andrew~M Dai, Rafal Jozefowicz,
  and Samy Bengio. 2016.
\newblock Generating sentences from a continuous space.
\newblock {\em COLING\/} .

\bibitem[{Callison-Burch et~al.(2011)Callison-Burch, Koehn, Monz, and
  Zaidan}]{callison2011findings}
Chris Callison-Burch, Philipp Koehn, Christof Monz, and Omar~F Zaidan. 2011.
\newblock Findings of the 2011 workshop on statistical machine translation.
\newblock In {\em Proceedings of the Sixth Workshop on Statistical Machine
  Translation\/}. Association for Computational Linguistics, pages 22--64.

\bibitem[{Chen and Cherry(2014)}]{chen2014systematic}
Boxing Chen and Colin Cherry. 2014.
\newblock A systematic comparison of smoothing techniques for sentence-level
  bleu.
\newblock {\em ACL 2014\/} page 362.

\bibitem[{Cohen(1968)}]{cohen1968weighted}
Jacob Cohen. 1968.
\newblock Weighted kappa: Nominal scale agreement provision for scaled
  disagreement or partial credit.
\newblock {\em Psychological bulletin\/} 70(4):213.

\bibitem[{Cooijmans et~al.(2016)Cooijmans, Ballas, Laurent, and
  Courville}]{cooijmans2016recurrent}
Tim Cooijmans, Nicolas Ballas, C{\'e}sar Laurent, and Aaron Courville. 2016.
\newblock Recurrent batch normalization.
\newblock {\em arXiv preprint arXiv:1603.09025\/} .

\bibitem[{DeVault et~al.(2011)DeVault, Leuski, and Sagae}]{devault2011toward}
David DeVault, Anton Leuski, and Kenji Sagae. 2011.
\newblock Toward learning and evaluation of dialogue policies with text
  examples.
\newblock In {\em Proceedings of the SIGDIAL 2011 Conference\/}. Association
  for Computational Linguistics, pages 39--48.

\bibitem[{Dhingra et~al.(2016)Dhingra, Zhou, Fitzpatrick, Muehl, and
  Cohen}]{dhingra2016tweet2vec}
Bhuwan Dhingra, Zhong Zhou, Dylan Fitzpatrick, Michael Muehl, and William~W
  Cohen. 2016.
\newblock Tweet2vec: Character-based distributed representations for social
  media.
\newblock {\em arXiv preprint arXiv:1605.03481\/} .

\bibitem[{Edwards and Storkey(2016)}]{edwards2015censoring}
Harrison Edwards and Amos Storkey. 2016.
\newblock Censoring representations with an adversary.
\newblock {\em ICLR\/} .

\bibitem[{El~Hihi and Bengio(1995)}]{el1995hierarchical}
Salah El~Hihi and Yoshua Bengio. 1995.
\newblock Hierarchical recurrent neural networks for long-term dependencies.
\newblock In {\em NIPS\/}. Citeseer, volume 400, page 409.

\bibitem[{Gage(1994)}]{gage1994new}
Philip Gage. 1994.
\newblock A new algorithm for data compression.
\newblock {\em The C Users Journal\/} 12(2):23--38.

\bibitem[{Galley et~al.(2015)Galley, Brockett, Sordoni, Ji, Auli, Quirk,
  Mitchell, Gao, and Dolan}]{galley2015deltableu}
Michel Galley, Chris Brockett, Alessandro Sordoni, Yangfeng Ji, Michael Auli,
  Chris Quirk, Margaret Mitchell, Jianfeng Gao, and Bill Dolan. 2015.
\newblock deltableu: A discriminative metric for generation tasks with
  intrinsically diverse targets.
\newblock {\em arXiv preprint arXiv:1506.06863\/} .

\bibitem[{Gandhe and Traum(2016)}]{gandhe2016semi}
Sudeep Gandhe and David Traum. 2016.
\newblock A semi-automated evaluation metric for dialogue model coherence.
\newblock In {\em Situated Dialog in Speech-Based Human-Computer
  Interaction\/}, Springer, pages 217--225.

\bibitem[{Gupta et~al.(2015)Gupta, Orasan, and van Genabith}]{gupta2015reval}
Rohit Gupta, Constantin Orasan, and Josef van Genabith. 2015.
\newblock Reval: A simple and effective machine translation evaluation metric
  based on recurrent neural networks.
\newblock In {\em Proceedings of the 2015 Conference on Empirical Methods in
  Natural Language Processing (EMNLP)\/}.

\bibitem[{Higashinaka et~al.(2014)Higashinaka, Meguro, Imamura, Sugiyama,
  Makino, and Matsuo}]{higashinaka2014evaluating}
Ryuichiro Higashinaka, Toyomi Meguro, Kenji Imamura, Hiroaki Sugiyama, Toshiro
  Makino, and Yoshihiro Matsuo. 2014.
\newblock Evaluating coherence in open domain conversational systems.
\newblock In {\em INTERSPEECH\/}. pages 130--134.

\bibitem[{Hochreiter(1991)}]{hochreiter1991untersuchungen}
Sepp Hochreiter. 1991.
\newblock Untersuchungen zu dynamischen neuronalen netzen.
\newblock {\em Diploma, Technische Universit{\"a}t M{\"u}nchen\/} page~91.

\bibitem[{Hochreiter and Schmidhuber(1997)}]{hochreiter1997long}
Sepp Hochreiter and J{\"u}rgen Schmidhuber. 1997.
\newblock Long short-term memory.
\newblock {\em Neural computation\/} 9(8):1735--1780.

\bibitem[{Ioffe and Szegedy(2015)}]{ioffe2015batch}
Sergey Ioffe and Christian Szegedy. 2015.
\newblock Batch normalization: Accelerating deep network training by reducing
  internal covariate shift.
\newblock {\em arXiv preprint arXiv:1502.03167\/} .

\bibitem[{Kannan et~al.(2016)Kannan, Kurach, Ravi, Kaufmann, Tomkins, Miklos,
  Corrado, Luk{\'a}cs, Ganea, Young et~al.}]{kannan2016smart}
Anjuli Kannan, Karol Kurach, Sujith Ravi, Tobias Kaufmann, Andrew Tomkins,
  Balint Miklos, Greg Corrado, L{\'a}szl{\'o} Luk{\'a}cs, Marina Ganea, Peter
  Young, et~al. 2016.
\newblock Smart reply: Automated response suggestion for email.
\newblock In {\em Proceedings of the ACM SIGKDD Conference on Knowledge
  Discovery and Data Mining (KDD)\/}. volume~36, pages 495--503.

\bibitem[{Kannan and Vinyals(2017)}]{kannan2017adversarial}
Anjuli Kannan and Oriol Vinyals. 2017.
\newblock Adversarial evaluation of dialogue models.
\newblock {\em arXiv preprint arXiv:1701.08198\/} .

\bibitem[{Kingma and Ba(2014)}]{kingma2014adam}
Diederik Kingma and Jimmy Ba. 2014.
\newblock Adam: A method for stochastic optimization.
\newblock {\em arXiv preprint arXiv:1412.6980\/} .

\bibitem[{Kiros et~al.(2015)Kiros, Zhu, Salakhutdinov, Zemel, Urtasun,
  Torralba, and Fidler}]{kiros2015skip}
Ryan Kiros, Yukun Zhu, Ruslan~R Salakhutdinov, Richard Zemel, Raquel Urtasun,
  Antonio Torralba, and Sanja Fidler. 2015.
\newblock Skip-thought vectors.
\newblock In {\em Advances in Neural Information Processing Systems\/}. pages
  3276--3284.

\bibitem[{Li et~al.(2015)Li, Galley, Brockett, Gao, and
  Dolan}]{li2015diversity}
Jiwei Li, Michel Galley, Chris Brockett, Jianfeng Gao, and Bill Dolan. 2015.
\newblock A diversity-promoting objective function for neural conversation
  models.
\newblock {\em arXiv preprint arXiv:1510.03055\/} .

\bibitem[{Li et~al.(2016{\natexlab{a}})Li, Galley, Brockett, Gao, and
  Dolan}]{li2016persona}
Jiwei Li, Michel Galley, Chris Brockett, Jianfeng Gao, and Bill Dolan.
  2016{\natexlab{a}}.
\newblock A persona-based neural conversation model.
\newblock {\em arXiv preprint arXiv:1603.06155\/} .

\bibitem[{Li et~al.(2017)Li, Monroe, and Jurafsky}]{li2017learning}
Jiwei Li, Will Monroe, and Dan Jurafsky. 2017.
\newblock Learning to decode for future success.
\newblock {\em arXiv preprint arXiv:1701.06549\/} .

\bibitem[{Li et~al.(2016{\natexlab{b}})Li, Monroe, Ritter, and
  Jurafsky}]{li2016deep}
Jiwei Li, Will Monroe, Alan Ritter, and Dan Jurafsky. 2016{\natexlab{b}}.
\newblock Deep reinforcement learning for dialogue generation.
\newblock {\em arXiv preprint arXiv:1606.01541\/} .

\bibitem[{Lin(2004)}]{lin2004rouge}
Chin-Yew Lin. 2004.
\newblock Rouge: A package for automatic evaluation of summaries.
\newblock In {\em Text summarization branches out: Proceedings of the ACL-04
  workshop\/}. Barcelona, Spain, volume~8.

\bibitem[{Liu et~al.(2016)Liu, Lowe, Serban, Noseworthy, Charlin, and
  Pineau}]{liu2016not}
Chia-Wei Liu, Ryan Lowe, Iulian~V Serban, Michael Noseworthy, Laurent Charlin,
  and Joelle Pineau. 2016.
\newblock How not to evaluate your dialogue system: An empirical study of
  unsupervised evaluation metrics for dialogue response generation.
\newblock {\em arXiv preprint arXiv:1603.08023\/} .

\bibitem[{Lowe et~al.(2015)Lowe, Pow, Serban, and Pineau}]{lowe2015ubuntu}
Ryan Lowe, Nissan Pow, Iulian Serban, and Joelle Pineau. 2015.
\newblock The ubuntu dialogue corpus: A large dataset for research in
  unstructured multi-turn dialogue systems.
\newblock {\em arXiv preprint arXiv:1506.08909\/} .

\bibitem[{Mach{\'a}cek and Bojar(2014)}]{machacek2014results}
Matou{\v{s}} Mach{\'a}cek and Ondrej Bojar. 2014.
\newblock Results of the wmt14 metrics shared task.
\newblock In {\em Proceedings of the Ninth Workshop on Statistical Machine
  Translation\/}. Citeseer, pages 293--301.

\bibitem[{Markoff and Mozur(2015)}]{markoff2015forsymp}
J.~Markoff and P.~Mozur. 2015.
\newblock For sympathetic ear, more chinese turn to smartphone program.
\newblock {\em NY Times\/} .

\bibitem[{M{\"o}ller et~al.(2006)M{\"o}ller, Englert, Engelbrecht, Hafner,
  Jameson, Oulasvirta, Raake, and Reithinger}]{moller2006memo}
Sebastian M{\"o}ller, Roman Englert, Klaus-Peter Engelbrecht, Verena~Vanessa
  Hafner, Anthony Jameson, Antti Oulasvirta, Alexander Raake, and Norbert
  Reithinger. 2006.
\newblock Memo: towards automatic usability evaluation of spoken dialogue
  services by user error simulations.
\newblock In {\em INTERSPEECH\/}.

\bibitem[{Papineni et~al.(2002)Papineni, Roukos, Ward, and
  Zhu}]{papineni2002bleu}
Kishore Papineni, Salim Roukos, Todd Ward, and Wei-Jing Zhu. 2002.
\newblock Bleu: a method for automatic evaluation of machine translation.
\newblock In {\em Proceedings of the 40th annual meeting on association for
  computational linguistics\/}. Association for Computational Linguistics,
  pages 311--318.

\bibitem[{Pearson(1901)}]{pearson1901principal}
Karl Pearson. 1901.
\newblock Principal components analysis.
\newblock {\em The London, Edinburgh and Dublin Philosophical Magazine and
  Journal\/} 6(2):566.

\bibitem[{Ritter et~al.(2011)Ritter, Cherry, and Dolan}]{ritter2011data}
Alan Ritter, Colin Cherry, and William~B Dolan. 2011.
\newblock Data-driven response generation in social media.
\newblock In {\em Proceedings of the conference on empirical methods in natural
  language processing\/}. Association for Computational Linguistics, pages
  583--593.

\bibitem[{Sennrich et~al.(2015)Sennrich, Haddow, and
  Birch}]{sennrich2015neural}
Rico Sennrich, Barry Haddow, and Alexandra Birch. 2015.
\newblock Neural machine translation of rare words with subword units.
\newblock {\em arXiv preprint arXiv:1508.07909\/} .

\bibitem[{Serban et~al.(2016{\natexlab{a}})Serban, Sordoni, Bengio, Courville,
  and Pineau}]{DBLP:conf/aaai/SerbanSBCP16}
Iulian~Vlad Serban, Alessandro Sordoni, Yoshua Bengio, Aaron Courville, and
  Joelle Pineau. 2016{\natexlab{a}}.
\newblock Building end-to-end dialogue systems using generative hierarchical
  neural network models.
\newblock In {\em AAAI\/}. pages 3776--3784.

\bibitem[{Serban et~al.(2016{\natexlab{b}})Serban, Sordoni, Lowe, Charlin,
  Pineau, Courville, and Bengio}]{serban2016hierarchical}
Iulian~Vlad Serban, Alessandro Sordoni, Ryan Lowe, Laurent Charlin, Joelle
  Pineau, Aaron Courville, and Yoshua Bengio. 2016{\natexlab{b}}.
\newblock A hierarchical latent variable encoder-decoder model for generating
  dialogues.
\newblock {\em arXiv preprint arXiv:1605.06069\/} .

\bibitem[{Shang et~al.(2015)Shang, Lu, and Li}]{shang2015neural}
Lifeng Shang, Zhengdong Lu, and Hang Li. 2015.
\newblock Neural responding machine for short-text conversation.
\newblock {\em arXiv preprint arXiv:1503.02364\/} .

\bibitem[{Shang et~al.(2016)Shang, Sakai, Lu, Li, Higashinaka, and
  Miyao}]{shang2016overview}
Lifeng Shang, Tetsuya Sakai, Zhengdong Lu, Hang Li, Ryuichiro Higashinaka, and
  Yusuke Miyao. 2016.
\newblock Overview of the ntcir-12 short text conversation task.
\newblock {\em Proceedings of NTCIR-12\/} pages 473--484.

\bibitem[{Sordoni et~al.(2015{\natexlab{a}})Sordoni, Bengio, Vahabi, Lioma,
  Grue~Simonsen, and Nie}]{sordoni2015hierarchical}
Alessandro Sordoni, Yoshua Bengio, Hossein Vahabi, Christina Lioma, Jakob
  Grue~Simonsen, and Jian-Yun Nie. 2015{\natexlab{a}}.
\newblock A hierarchical recurrent encoder-decoder for generative context-aware
  query suggestion.
\newblock In {\em Proceedings of the 24th ACM International on Conference on
  Information and Knowledge Management\/}. ACM, pages 553--562.

\bibitem[{Sordoni et~al.(2015{\natexlab{b}})Sordoni, Galley, Auli, Brockett,
  Ji, Mitchell, Nie, Gao, and Dolan}]{sordoni2015neural}
Alessandro Sordoni, Michel Galley, Michael Auli, Chris Brockett, Yangfeng Ji,
  Margaret Mitchell, Jian-Yun Nie, Jianfeng Gao, and Bill Dolan.
  2015{\natexlab{b}}.
\newblock A neural network approach to context-sensitive generation of
  conversational responses.
\newblock {\em arXiv preprint arXiv:1506.06714\/} .

\bibitem[{Stanojevic et~al.(2015)Stanojevic, Kamran, Koehn, and
  Bojar}]{stanojevic2015results}
Milo{\v{s}} Stanojevic, Amir Kamran, Philipp Koehn, and Ondrej Bojar. 2015.
\newblock Results of the wmt15 metrics shared task.
\newblock In {\em Proceedings of the Tenth Workshop on Statistical Machine
  Translation\/}. pages 256--273.

\bibitem[{Turing(1950)}]{turing1950computing}
Alan~M Turing. 1950.
\newblock Computing machinery and intelligence.
\newblock {\em Mind\/} 59(236):433--460.

\bibitem[{Vinyals and Le(2015)}]{vinyals2015neural}
Oriol Vinyals and Quoc Le. 2015.
\newblock A neural conversational model.
\newblock {\em arXiv preprint arXiv:1506.05869\/} .

\bibitem[{Walker et~al.(1997)Walker, Litman, Kamm, and
  Abella}]{walker1997paradise}
Marilyn~A Walker, Diane~J Litman, Candace~A Kamm, and Alicia Abella. 1997.
\newblock Paradise: A framework for evaluating spoken dialogue agents.
\newblock In {\em Proceedings of the eighth conference on European chapter of
  the Association for Computational Linguistics\/}. Association for
  Computational Linguistics, pages 271--280.

\bibitem[{Weizenbaum(1966)}]{weizenbaum1966eliza}
J.~Weizenbaum. 1966.
\newblock {ELIZA}—a computer program for the study of natural language
  communication between man and machine.
\newblock {\em Communications of the ACM\/} 9(1):36--45.

\bibitem[{Xiang et~al.(2014)Xiang, Zhang, Zhou, Wang, and
  Qin}]{xiang2014problematic}
Yang Xiang, Yaoyun Zhang, Xiaoqiang Zhou, Xiaolong Wang, and Yang Qin. 2014.
\newblock Problematic situation analysis and automatic recognition for chi-nese
  online conversational system.
\newblock {\em Proc. CLP\/} pages 43--51.

\bibitem[{Yu et~al.(2016)Yu, Xu, Black, and Rudnicky}]{yu2016strategy}
Zhou Yu, Ziyu Xu, Alan~W Black, and Alex~I Rudnicky. 2016.
\newblock Strategy and policy learning for non-task-oriented conversational
  systems.
\newblock In {\em 17th Annual Meeting of the Special Interest Group on
  Discourse and Dialogue\/}. page 404.

\end{thebibliography}
\bibliographystyle{acl_natbib}

\newpage

\section*{Appendix A: Further Notes on Crowdsourcing Data Collection}

\paragraph{Amazon Mechanical Turk Experiments}

We conducted two rounds of AMT experiments.
We first asked AMT workers to provide a reasonable continuation of a Twitter dialogue (i.e.\@ generate the next response given the context of a conversation).
Each survey contained 20 questions, including an attention check question.
Workers were instructed to generate longer responses, in order to avoid simple one-word responses.
In total, we obtained approximately 2,000 human responses.

Second, we filtered these human-generated responses for potentially offensive language, and combined them with approximately 1,000 responses from each of the above models into a single set of responses. 
We then asked AMT workers to rate the overall quality of each response on a scale of 1 (low quality) to 5 (high quality). Each user was asked to evaluate 4 responses from 50 different contexts. 
We included four additional attention-check questions and a set of five contexts was given to each participant for assessment of inter-annotator agreement. We removed all users who either failed an attention check question or achieved a $\kappa$ inter-annotator agreement score lower than 0.2~\citep{cohen1968weighted}.
The remaining evaluators had a median $\kappa$ score of 0.63, indicating moderate agreement.
This is consistent with results from \cite{liu2016not}. 
Dataset statistics are provided in Table \ref{tab:dataset}.

In initial experiments, we also asked humans to provide scores for topicality, informativeness, and whether the context required background information to be understandable. Note that we did not ask for fluency scores, as 3/4 of the responses were produced by humans (including the retrieval models).
We found that scores for informativeness and background had low inter-annotator agreement (Table \ref{tab:kappa}), and scores for topicality were highly correlated with the overall score (Pearson correlation of 0.72).
Results on these auxiliary questions varied depending on the wording of the question. Thus, we continued our experiments by only asking for the overall score. 
We provide more details concerning the data collection in the supplemental material, as it may aid others in developing effective crowdsourcing experiments. 

\begin{table}
    \centering
    \begin{tabular}{l r}
         \toprule
          \textbf{Measurement }& \textbf{$\kappa$ score }\\ \hline
          Overall & 0.63 \\
         Topicality & 0.57 \\ 
         Informativeness & 0.31  \\ 
         Background & 0.05 \\ \bottomrule       
    \end{tabular}
    \caption{Median $\kappa$ inter-annotator agreement scores for various questions asked in the survey. 
    }
    \label{tab:kappa}
    \vspace{-4mm}
\end{table}

\paragraph{Preliminary AMT experiments}

Before conducting the primary crowdsourcing experiments to collect the dataset in this paper, we ran a series of preliminary experiments to see how AMT workers responded to different questions. Unlike the primary study, where we asked a small number of overlapping questions to determine the $\kappa$ score and filtered users based on the results, we conducted a study where all responses (40 in total from 10 contexts) were overlapping. We did this for 18 users in two trials, resulting in 153 pair-wise correlation scores per trial.

In the first trial, we asked the following questions to the users, for each response:
\begin{enumerate}
    \item How appropriate is the response overall? (overall, scale of 1-5)
    \item How on-topic is the response? (topicality, scale of 1-5)
    \item How specific is the response to some context? (specificity, scale of 1-5)
    \item How much background information is required to understand the context? (background, scale of 1-5)
\end{enumerate}
Note that we do not ask for fluency, as the 3/4 responses for each context were written by a human (including retrieval models). We also provided the AMT workers with examples that have high topicality and low specificity, and examples with high specificity and low topicality. The background question was only asked once for each context.

We observed that both the overall scores and topicality had fairly high inter-annotator agreement (as shown in Table \ref{tab:kappa}), but were strongly correlated with each other (i.e.\@ participants would often put the same scores for topicality and overall score). Conversely, specificity ($\kappa = 0.12$) and background ($\kappa = 0.05$) had very low inter-annotator agreements. 

\begin{figure*}
\centering
\includegraphics[width=.9\linewidth]{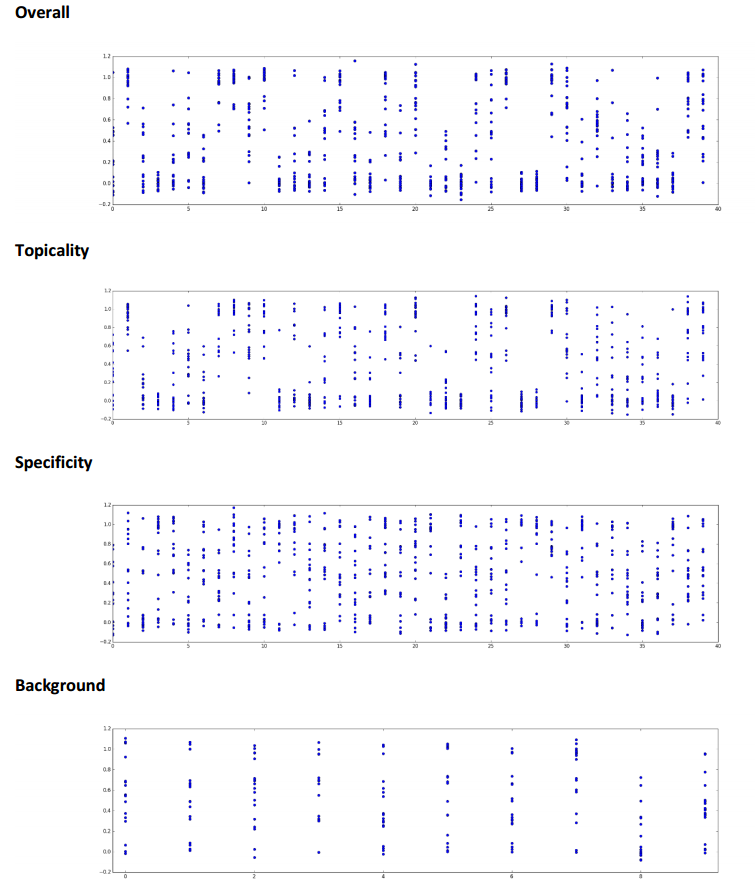}
\caption{\label{fig:scoreplot} Scatter plots showing the distribution of scores (vertical axis) for different responses (horizontal axis), for each of the four questions in our survey. It can be seen that the overall and topicality scores are clustered for each question, indicating high agreement, while this is not the case for specificity or background information. Note that all scores are normalized based on a per-user basis, based on the average score given by each user.}
\end{figure*}

To better visualize the data, we produce scatterplots showing the distribution of scores for different responses, for each of the four questions in our survey (Figure \ref{fig:scoreplot}). We can see that the overall and topicality scores are clustered for each question, indicating high agreement.  However, these clusters are most often in the same positions for each response, which indicates that they are highly correlated with each other. Specificity and background information, on the other hand, show far fewer clusters, indicating lower inter-annotator agreement. We conjectured that this was partially because the terms `specificity' and `background information', along with our descriptions of them, had a high cognitive load, and were difficult to understand in the context of our survey.

To test this hypothesis, we conducted a new survey where we tried to ask the questions for specificity and background in a more intuitive manner. We also changed the formulation of the background question to be a binary 0-1 decision of whether users understood the context. We asked the following questions:
\begin{enumerate}
    \item How appropriate is the response overall? (overall, scale of 1-5)
    \item How on-topic is the response? (topicality, scale of 1-5)
    \item How common is the response? (informativeness, scale of 1-5)
    \item Does the context make sense? (context, scale of 0-1)
\end{enumerate}
We also clarified our description for the third question, including providing more intuitive examples. Interestingly, the inter-annotator agreement on informativeness $\kappa=0.31$ was much higher than that for specificity in the original survey. Thus, the formulation of questions in a crowdsourcing survey has a large impact on inter-annotator agreement. For the context, we found that users either agreed highly ($\kappa>0.9$ for 45 participants), or not at all ($\kappa<0.1$ for 113 participants). 

We also experimented with asking the overall score on a separate page, before asking questions 2-4, and found that this increased the $\kappa$ agreement slightly. Similarly, excluding all scores where participants indicated they did not understand the context improved inter-annotator agreement slightly.

Due to these observations, we decided to only ask users for their overall quality score for each response, as it is unclear how much additional information is provided by the other questions in the context of dialogue. We hope this information is useful for future crowdsourcing experiments in the dialogue domain.

\section*{Appendix B: Metric Description}


\paragraph{BLEU}
BLEU~\citep{papineni2002bleu} analyzes the co-occurrences of n-grams in the ground truth and the proposed responses. It first computes an n-gram precision for the whole dataset:
$$
P_n(r,\hat{r}) = \frac{\sum_k \min(h(k,r), h(k,\hat{r}_i))}{\sum_k h(k,r_i)}
$$
where $k$ indexes all possible n-grams of length $n$ and $h(k,r)$ is the number of n-grams $k$ in $r$. Note that the min in this equation is calculating the number of co-occurrences of n-gram $k$ between the ground truth response $r$ and the proposed response $\hat{r}$, as it computes the fewest appearances of $k$ in either response. To avoid the drawbacks of using a precision score, namely that it favours shorter (candidate) sentences, the authors introduce a brevity penalty. BLEU-N, where $N$ is the maximum length of n-grams considered, is defined as:
$$
\text{\small BLEU-N} := b(r,\hat{r}) \exp (\sum^N_{n=1} \beta_n \log P_n(r,\hat{r}))
$$
 $\beta_n$ is a weighting that is usually uniform, and $b(\cdot)$ is the brevity penalty. The most commonly used version of BLEU assigns $N=4$. Modern versions of BLEU also use sentence-level smoothing, as the geometric mean often results in scores of 0 if there is no 4-gram overlap~\citep{chen2014systematic}. Note that BLEU is usually calculated at the corpus-level, and was originally designed for use with multiple reference sentences.



\paragraph{METEOR}

The METEOR metric \citep{banerjee2005meteor} was introduced to address several weaknesses in BLEU. It creates an explicit alignment between the candidate and target responses. The alignment is based on exact token matching, followed by WordNet synonyms, stemmed tokens, and then paraphrases. Given a set of alignments, the METEOR score is the harmonic mean of precision and recall between the proposed and ground truth sentence.

Given a set of alignments $m$, the METEOR score is the harmonic mean of precision $P_m$ and recall $R_m$ between the candidate and target sentence.
\begin{gather} 
Pen = \gamma (\frac{ch}{m})^\theta \\
F_{mean} = \frac{P_m R_m}{\alpha P_m + (1 - \alpha) R_m} \\
P_m = \frac{|m|}{\sum_k h_k(c_i)} \\
R_m = \frac{|m|}{\sum_k h_k(s_{ij})} \\
METEOR = (1 - Pen)F_{mean}
\end{gather}
The penalty term $Pen$ is based on the `chunkiness' of the resolved matches. We use the default values for the hyperparameters $\alpha, \gamma$, and $\theta$.


\paragraph{ROUGE} ROUGE~\citep{lin2004rouge} is a set of evaluation metrics used for automatic summarization. We consider ROUGE-L, which is a F-measure based on the Longest Common Subsequence (LCS) between a candidate and target sentence. The LCS is a set of words which occur in two sentences in the same order; however, unlike n-grams the words do not have to be contiguous, i.e.\@ there can be other words in between the words of the LCS. ROUGE-L is computed using an F-measure between the reference response and the proposed response. 
\begin{gather}
R = \max_j \frac{l(c_i, s_{ij})}{|s_{ij}|} \\
P = \max_j frac{l(c_i, s_{ij})}{|c_{ij}|} \\
ROUGE_L(c_i, S_i) = \frac{(1 + \beta^2)RP}{R + \beta^2 P}
\end{gather}
where $l(c_i, s_{ij})$ is the length of the LCS between the sentences. $\beta$ is usually set to favour recall ($\beta=1.2$). 

\section*{Appendix C: Latent Variable Hierarchical Recurrent Encoder-Decoder (VHRED)}

The VHRED model is an extension of the original hierarchical recurrent encoder-decoder (HRED) model~\citep{DBLP:conf/aaai/SerbanSBCP16} with an additional component: a high-dimensional stochastic latent variable at every dialogue turn. 
The dialogue context is encoded into a vector representation using the \textit{utterance-level} and \textit{context-level} RNNs from our encoder. Conditioned on the summary vector at each dialogue turn, 
VHRED samples a multivariate Gaussian variable that is provided, along with the context summary vector, as input to the \textit{decoder} RNN, which in turn generates the response word-by-word. We use representations from the VHRED model as it produces more diverse and coherent responses compared to its HRED counterpart.

The VHRED model is trained to maximize a lower-bound on the log-likelihood of generating the next response:
\begin{align}
\mathcal{L} &= \log P_{\hat{\theta}}(\mathbf{w}_1, \dots, \mathbf{w}_N) \nonumber \\ 
& \geq  \sum_{n=1}^N - \text{KL} \left [ Q_{\psi}(\mathbf{z}_n \mid \mathbf{w}_1, \dots, \mathbf{w}_n) || P_{\hat{\theta}}(\mathbf{z}_n \mid \mathbf{w}_{<n} ) \right ] \nonumber \\ &+ \mathbb{E}_{Q_{\psi}(\mathbf{z}_n \mid \mathbf{w}_1, \dots, \mathbf{w}_n)} \left [ \log P_{\hat{\theta}}(\mathbf{w}_n \mid \mathbf{z}_n, \mathbf{w}_{< n}) \right ], \label{VHRED:lower_bound}
\end{align}
where $\text{KL}[Q || P]$ is the Kullback-Leibler (KL) divergence between distributions $Q$ and $P$. The distribution $Q_{\psi}(\mathbf{z}_n \mid \mathbf{w}_1, \dots, \mathbf{w}_N) =  \mathcal{N}(\boldsymbol{\mu}_{\text{posterior}}(\mathbf{w}_1, \dots, \mathbf{w}_n), \Sigma_{\text{posterior}}(\mathbf{w}_1, \dots, \mathbf{w}_n))$ is the approximate posterior distribution (or \textit{recognition model}) which approximates the intractable true posterior distribution $P_{\psi}(\mathbf{z}_n \mid \mathbf{w}_1, \dots, \mathbf{w}_N)$. The posterior mean $\boldsymbol{\mu}_{\text{posterior}}$ and covariance $\Sigma_{\text{posterior}}$ (as well as that of the prior) are computed using a feed-forward neural network, which takes as input the concatenation of the vector representations of the past utterances and that of the current utterance.

The multivariate Gaussian latent variable in the VHRED model allows modelling ambiguity and uncertainty in the dialogue through the latent variable distribution parameters (mean and variance).
This provides a useful inductive bias, which helps VHRED encode the dialogue context into a real-valued embedding space even when the dialogue context is ambiguous or uncertain, and it helps VHRED generate more diverse responses.

\paragraph{Pre-training motivation} Maximizing the likelihood of generating the next utterance in a dialogue is not only a convenient way of training the encoder parameters; it is also an objective that is consistent with learning useful representations of the dialogue utterances.
Two context vectors produced by the VHRED encoder are similar if the contexts induce a similar distribution over subsequent responses; this is consistent with the formulation of the evaluation model, which assigns high scores to responses that have similar vector representations to the context.
VHRED is also closely related to the skip-thought-vector model~\citep{kiros2015skip}, which has been shown to learn useful representations of sentences for many tasks, including semantic relatedness and paraphrase detection.
The skip-thought-vector model takes as input a single sentence and predicts the previous sentence and next sentence.
On the other hand, VHRED takes as input several consecutive sentences and predicts the next sentence.
This makes it particularly suitable for learning long-term context representations.


\section*{Appendix D: Experiments \& results}

\subsection*{Hyperparameters}
When evaluating our model, we conduct early stopping on an external validation set to obtain the best parameter setting. We similarly choose our  hyperparameters (PCA dimension $n$, L2 regularization penalty $\gamma$, learning rate $a$, and batch size $b$) based on validation set results. Our best {\sc adem} model used $\gamma=0.075$, $a = 0.01$, and $b = 32$. 
For {\sc adem} with tweet2vec embeddings, we did a similar hyperparameter searched, and used $n=150$, $\gamma=0.01$, $a = 0.01$, and $b = 16$.

\subsection*{Additional Results}

\begin{table}
\footnotesize
    \centering
    \begin{tabular}{l c c}
         \toprule                    
         \textbf{Metric} & \textbf{Spearman} & \textbf{Pearson} \\ \hline
         BLEU-1 & -0.026 (0.80) & 0.016 (0.87) \\
         BLEU-2 & 0.065 (0.52) & 0.080 (0.43) \\
         BLEU-3 & 0.139 (0.17) & 0.088 (0.39) \\
         BLEU-4 & 0.139 (0.17) & 0.092 (0.36) \\
         ROUGE & -0.083 (0.41) & -0.010 (0.92) \\         
           \bottomrule
    \end{tabular}
    \caption{Correlations between word-overlap metrics and human judgements on the dataset from \cite{liu2016not}, after removing the speaker tokens at the beginning of each utterance. The correlations are even worse than estimated in the original paper, and none are significant. }
    \label{tab:liu}
    \vspace{-3mm}
\end{table}
\paragraph{New results on \cite{liu2016not} data } In order to ensure that the correlations between word-overlap metrics and human judgements were comparable across datasets, we standardized the processing of the evaluation dataset from \cite{liu2016not}. In particular, the original data from \cite{liu2016not} has a token (either `\<first\_speaker\>', `\<second\_speaker\>', or `\<third\_speaker\>') at the beginning of each utterance. This is an artifact left-over by the processing used as input to the hierarchical recurrent encoder-decoder (HRED) model \citep{DBLP:conf/aaai/SerbanSBCP16}. Removing these tokens makes sense for establishing the ability of word-overlap models, as they are unrelated to the content of the tweets.

We perform this processing, and report the updated results for word-overlap metrics in Table \ref{tab:liu}. Surprisingly, almost all significant correlation disappears, particularly for all forms of the BLEU score. Thus, we can conclude that the word-overlap metrics were heavily relying on these tokens to form bigram matches between the model responses and reference responses.

\begin{table}
\footnotesize
    \centering
    \begin{tabular}{l c}
         \toprule          
         \textbf{Metric} & \textbf{Wall time} \\ \hline
         {\sc adem} (CPU) & 2861s \\
         {\sc adem} (GPU) & 168s \\         
           \bottomrule
    \end{tabular}
    \caption{Evaluation time on the test set.}
    \label{tab:speed}
\end{table}
\paragraph{Evaluation speed } An important property of evaluation models is speed.
We show the evaluation time on the test set for {\sc adem} on both CPU and a Titan X GPU (using Theano, without cudNN) in Table \ref{tab:speed}. 
When run on GPU, {\sc adem} is able to evaluate responses in a reasonable amount of time (approximately 2.5 minutes). This includes the time for encoding the contexts, model responses, and reference responses into vectors with the hierarchical RNN, in addition to computing the PCA projection, but does not include pre-training with VHRED. 
For comparison, if run on a test set of 10,000 responses, {\sc adem} would take approximately 45 minutes. This is significantly less time consuming than setting up human experiments at any scale. Note that we have not yet made any effort to optimize the speed of the {\sc adem} model.

\begin{figure*}[h]
\centering
\begin{subfigure}[b]{0.4\textwidth}
            \centering
            \includegraphics[width=.95\linewidth]{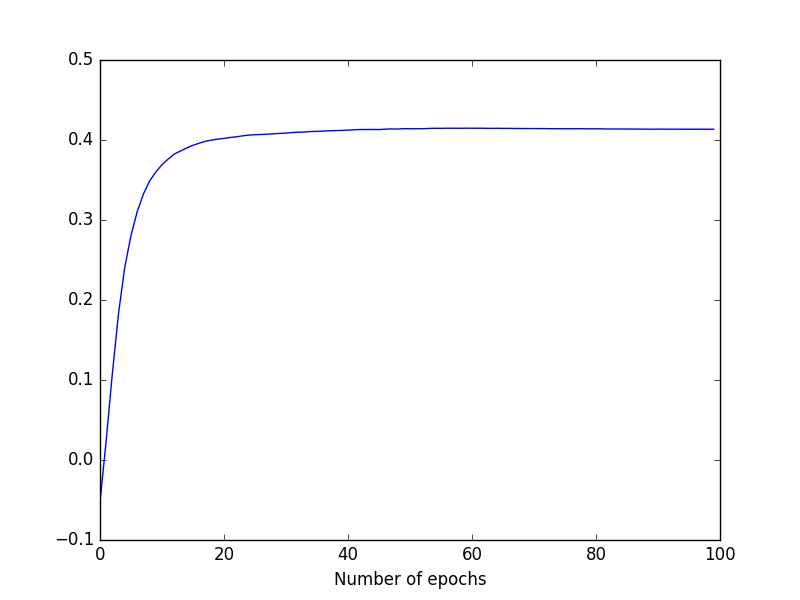}
            \caption{Spearman correlation}
\end{subfigure}
\begin{subfigure}[b]{0.4\textwidth}
            \centering
            \includegraphics[width=.95\linewidth]{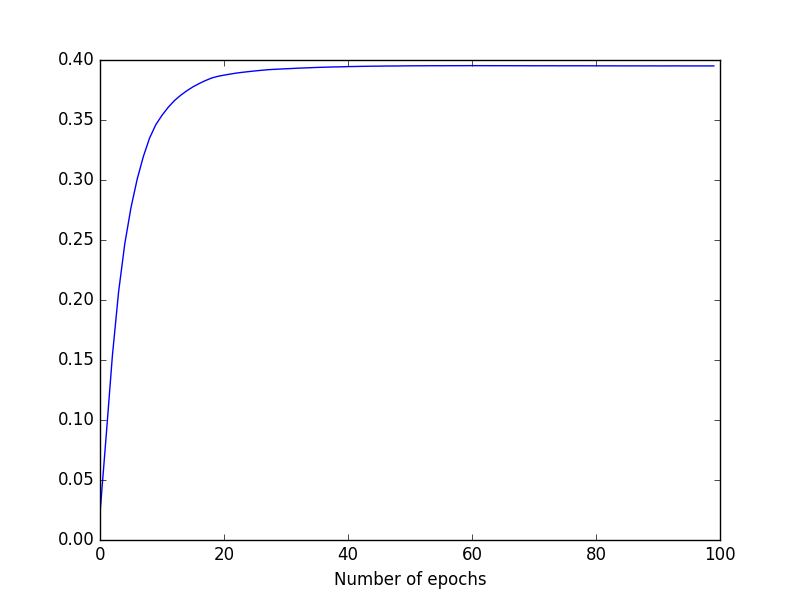}
            \caption{Pearson correlation}
\end{subfigure}\\
\vspace{-2mm}
\caption{\label{fig:learningcurves} Plots showing the Spearman and Pearson correlations on the test set as {\sc adem} trains. At the beginning of training, the model does not correlate with human judgements.   }
\end{figure*}

\paragraph{Learning curves } To show that our learning procedure for {\sc adem} really is necessary, and that the embeddings produced by VHRED are not sufficient to evaluate dialogue systems, we plot the Spearman and Pearson correlations on the test set as a function of the number of epochs in Figure \ref{fig:learningcurves}. It is clear that, at the beginning of training, when the matrices $M$ and $N$ have been initialized to the identity, the model is incapable of accurately predicting human scores, and its correlation is approximately 0.

\newpage

\paragraph{Failure analysis } We now conduct a failure analysis of the {\sc adem} model. In particular, we look at two different cases: responses where both humans and (normalized) ROUGE or BLEU-2 score highly (a score of 4 out of 5 or greater) while {\sc adem} scores poorly (2 out of 5 or lower), and the converse, where {\sc adem} scores the response highly while humans and either ROUGE or BLEU-2 score it poorly. We randomly sample (i.e.\@ without cherry picking) three examples of each case, which are shown in Tables \ref{tab:failure1}-\ref{tab:failure2}.

From Table \ref{tab:failure1}, the cases where {\sc adem} misses a good response, we can see that there are a variety of reasons for this cause of failure. In the first example, {\sc adem} is not able to match the fact that the model response talks about sleep to the reference response or context. This is possibly because the utterance contains a significant amount of irrelevant information: indeed, the first two sentences are not related to either the context or reference response. In the second example, the model response does not seem particularly relevant to the context --- despite this, the human scoring this example gave it $4/5$. This illustrates one drawback of human evaluations; they are quite subjective, and often have some noise. This makes it difficult to learn an effective {\sc adem} model. Finally, {\sc adem} is unable to score the third response highly, even though it is very closely related to the reference response. 

We can observe from the first two examples in Table \ref{tab:failure2}, where the {\sc adem} model erroneously ranks the model responses highly, that {\sc adem} is occasionally fooled into giving high scores for responses that are completely unrelated to the context. This may be because both of the utterances are short, and short utterances are ranked higher by humans in general since they are often more generic (as detailed in Section \ref{sec:experiments}). In the third example, the response actually seems to be somewhat reasonable given the context; this may be an instance where the human evaluator provided a score that was too low.

\begin{table*}
\small
\centering
\begin{tabular}{p{45mm} p{17mm} p{18mm} p{8mm} p{8mm} p{8mm} p{8mm}}
\toprule
\textbf{Context} & \textbf{Reference response} & \textbf{Model response} & \textbf{Human score} & \textbf{$|$BLEU$\text{-}$2$|$ score} & \textbf{$|$ROUGE$|$ score} & \textbf{$|${\sc adem}$|$ score} \\ \hline

what theme do you guys want next on tumblr? we've had mariskamommymoments what do you want to see next? $\rightarrow$  { \color{blue} im sorry. hope you feel better soon!} $\rightarrow$ it will wear off. just hate feeling like this $\rightarrow$  { \color{blue} im sure it will! just relax and take your time} $\rightarrow$ i'm okay. just overly tired
 & maybe you need a bit more sleep 
 & i'm really impressed. first guy to said that p what's time? sleep late its not good. i'm worried
 & 4
 & 2.53
 & 5.0
 & 1.0
 \\ \hline
 1 some pm syria - the editor of syrian government daily tishrin said she had been sacked over remarks to al  \<url\> $\rightarrow$  { \color{blue} replaced by business editor! sounds like the states lol}
 & msm is very similar the world over ! tied to the government .
 & i'm not sure if i'm going to be able to see the \<unk\>. i'm not sure if i should be able to see it.
 & 4
 & 2.53
 & 4.75
 & 1.22
 \\ \hline
 wonder how long it will take wayne rooney to get himself into trouble on twitter. odds? $\rightarrow$  { \color{blue} probably. a long time. because i used the address in the paper to find the page and it doesn't exist!} $\rightarrow$ here you go 
 & thanks . i wonder why it didn't work !
 & thanks let me see if this one works :p
 & 5
 & 2.53
 & 4.24
 & 1.53 \\
\bottomrule
\end{tabular}
\vspace{-1mm}
\caption{\label{tab:failure1}  Examples where a human and either BLEU-2 or ROUGE (after normalization) score the model response highly ($>4/5$), while the {\sc adem} model scored it poorly ($<2/5$). These examples are drawn randomly (i.e.\@ no cherry-picking). The bars around $|$metric$|$ indicate that the metric scores have been normalized. }
\vspace{-4mm}
\end{table*}

\begin{table*}
\small
\centering
\begin{tabular}{p{45mm} p{20mm} p{15mm} p{8mm} p{8mm} p{8mm} p{8mm}}
\toprule
\textbf{Context} & \textbf{Reference response} & \textbf{Model response} & \textbf{Human score} & \textbf{$|$BLEU$\text{-}$2$|$ score} & \textbf{$|$ROUGE$|$ score} & \textbf{$|${\sc adem}$|$ score} \\ \hline

rage slightly dissipated. wouldn't have bothered restoring my phone but i need it to moan at tomorrow. $\rightarrow$  { \color{blue} speaking of moaning. i'm actually going to email that chap that letter right now.} $\rightarrow$ good plan
 & i 'm going to do a little wee blog about it too . all nice and measured , of course .
 & some. some unfortunately
 & 2
 & 2.53
 & 1.57
 & 4.38
 \\ \hline
 high school flings college relationships $\rightarrow$  { \color{blue} it seems like the other way around from wat i've seen}
 & word . i 've seen a little of both . more of the college though
 & king james
 & 1
 & 2.53
 & 1.57
 & 5.0
 \\ \hline
 is it getting light outside? i swear it looks blue. $\rightarrow$  { \color{blue} time to go to sleepppp..}
 & for you , i 'm staying up
 & i'm going to the beach.
 & 1
 & 2.53
 & 1.57
 & 5.0 \\
\bottomrule
\end{tabular}
\vspace{-1mm}
\caption{\label{tab:failure2}  Examples where a human and either BLEU-2 or ROUGE (after normalization) score the model response low ($<2/5$), while the {\sc adem} model scored it highly ($>4/5$). These examples are drawn randomly (i.e.\@ no cherry-picking). The bars around $|$metric$|$ indicate that the metric scores have been normalized. }
\vspace{-4mm}
\end{table*}

\paragraph{Data efficiency } How much data is required to train {\sc adem}? We conduct an experiment where we train {\sc adem} on different amounts of training data, from 5\% to 100\%. The results are shown in Table \ref{tab:missingdata}. We can observe that {\sc adem} is very data-efficient, and is capable of reaching a Spearman correlation of 0.4 using only half of the available training data (1000 labelled examples). {\sc adem} correlates significantly with humans even when only trained on 5\% of the original training data (100 labelled examples).

\begin{table*}[h]
    \centering
    \begin{tabular}{c c c c c}
         \toprule          
         \textbf{Training data \%} & \textbf{Spearman} & p-value & \textbf{Pearson} & p-value \\ \hline
            100 \% of data & 0.414 & \< 0.001 & 0.395 & \< 0.001 \\       
            75 \% of data & 0.408 & \< 0.001 & 0.393 & \< 0.001 \\
            50 \% of data & 0.400 & \< 0.001 & 0.391 & \< 0.001 \\
            25 \% of data & 0.330 & \< 0.001 & 0.331 & \< 0.001 \\
            10 \% of data & 0.245 & \< 0.001 & 0.265 & \< 0.001 \\
            5 \% of data & 0.098 & 0.015 & 0.161 & \< 0.001 \\
           \bottomrule
    \end{tabular}
    \caption{{\sc adem} correlations when trained on different amounts of data.}
    \label{tab:missingdata}
\end{table*}

\paragraph{Improvement over word-overlap metrics } Next, we analyze more precisely how {\sc adem} outperforms traditional word-overlap metrics such as BLEU-2 and ROUGE. We first normalize the metric scores to have the same mean and variance as human scores, clipping the resulting scores to the range $[1,5]$ (we assign raw scores of 0 a normalized score of 1). \textit{We indicate normalization with vertical bars around the metric.} We then select all of the good responses that were given low scores by word-overlap metrics (i.e.\@ responses which humans scored as 4 or higher, and which $|$BLEU-2$|$ and $|$ROUGE$|$ scored as 2 or lower). The results are summarized in Table \ref{tab:numexamples}: of the 237 responses that humans scored 4 or higher, most of them (147/237) were ranked very poorly by both BLEU-2 and ROUGE. 
This quantitatively demonstrates what we argued qualitatively in Figure \ref{tab:toy}; a major failure of word-overlap metrics is the inability to consider
reasonable responses that have no word-overlap with the reference response. We can also see that, in almost half (60/147) of the cases where both BLEU-2 and ROUGE fail, $|${\sc adem}$|$ is able to correctly assign a score greater than 4. For comparison, there are only 42 responses where humans give a score of 4 and $|${\sc adem}$|$ gives a score less than 2, and only 14 of these are assigned a score greater than 4 by either $|$BLEU-2$|$ or $|$ROUGE$|$.

\begin{table}
\vspace{-3mm}
     \footnotesize
    \centering
        \begin{tabular}{l c}
            \toprule    
            \textbf{Metric scores} & \textbf{\# Examples} \\ \hline
            Human $\geq 4$  &  237 out of 616 \\ \hline
            \textbf{and} ($|$BLEU-2$|$ \<2, & \multirow{2}{*}{146 out of 237}  \\
            \MyInd\MyInd\MyInd  $|$ROUGE$|$ \<2) &   \vspace{.7mm}   \\        
            \textbf{and} $|${\sc adem}$|$ \> 4 &  60 out of 146   \\  \hline 
            \textbf{and} $|${\sc adem}$|$ \< 2 &  42 out of 237   \vspace{.9mm}\\ 
            \textbf{and} ($|$BLEU-2$|$ \>4, & \multirow{2}{*}{14 out of 42}  \\
            \MyInd\MyInd\MyInd \textbf{or}  $|$ROUGE$|$ \>4) &     \\        
            
           \bottomrule         
\end{tabular}
\caption{\label{tab:numexamples}In 60/146 cases, {\sc adem} scores good responses (human score \> 4) highly when word-overlap metrics fail. The bars around $|$metric$|$ indicate that the metric scores have been normalized. }
\vspace{-3mm}
\end{table}

\begin{table}
     \footnotesize
    \centering
        \begin{tabular}{l c c c}
            \toprule    
                & \multicolumn{2}{c}{\textbf{Mean score}} & \\
                & $\Delta w \leq 6$ & $\Delta w > 6$ & \textbf{p-value} \\
                & (n=312) & (n=304) & \\ \hline 
            ROUGE &  0.042   & 0.031  & \< 0.01  \\
            BLEU-2 & 0.0022   & 0.0007   & 0.23  \\
            {\sc adem} &    2.072 &  2.015 & 0.23 \\ \hline 
            Human &     2.671 &     2.698  & 0.83 \\           
                      
           \bottomrule         
\end{tabular}
\vspace{-2mm}
\caption{\label{tab:delw}Effect of differences in response length on the score, $\Delta w$ = absolute difference in \#words between the reference response and proposed response. BLEU-1, BLEU-2, and METEOR have previously been shown to exhibit bias towards similar-length responses~\citep{liu2016not}.}
\vspace{-4mm}
\end{table}
To provide further insight, we give specific examples of responses that are scored highly (\> 4) by both humans and $|${\sc adem}$|$, and poorly (\< 2) by both $|$BLEU-2$|$ and $|$ROUGE$|$ in Table \ref{tab:qualitative2}. We draw 3 responses randomly (i.e.\@ no cherry-picking) from the 60 test set responses that meet this criteria. We can observe that {\sc adem} is able to recognize short responses that are appropriate to the context, without word-overlap with the reference response. This is even the case when the model and reference responses have very little semantic similarity, as in the first and third examples in Table \ref{tab:qualitative2}.

\begin{table*}
\small
\centering
\begin{tabular}{p{40mm} p{30mm} p{10mm} p{8mm} p{8mm} p{8mm} p{8mm}}
\toprule
\textbf{Context} & \textbf{Reference response} & \textbf{Model response} & \textbf{Human score} & \textbf{$|$BLEU$\text{-}$2$|$ score} & \textbf{$|$ROUGE$|$ score} & \textbf{$|${\sc adem}$|$ score} \\ \hline
 i'd recommend \<url\> - or build buy an htpc and put \<url\> on it. $\rightarrow$  { \color{blue} you're the some nd person this week that's recommended roku to me.}
 & an htpc with xmbc is what i run . but i 've decked out my setup . i 've got \<number\> tb of data on my home server
 & because it's brilliant
 & 5
 & 1.0
 & 1.0
 & 4.726
 \\ \hline
 imma be an auntie this weekend. i guess i have to go albany. herewego $\rightarrow$  { \color{blue} u supposed to been here} $\rightarrow$ i come off nd on. $\rightarrow$  { \color{blue} never tell me smh}
 & lol you sometiming
 & haha, anyway, how're you?
 & 5
 & 1.0
 & 1.0
 & 4.201
 \\ \hline
 my son thinks she is plain. and the girl that plays her sister. seekhelp4him? $\rightarrow$  { \color{blue} send him this. he'll thank you. \<url\>}
 & you are too kind for words .
 & i will do
 & 5
 & 1.0
 & 1.0
 & 5.0 \\
\bottomrule
\end{tabular}
\vspace{-1mm}
\caption{\label{tab:qualitative2}  Examples where both human and {\sc adem} score the model response highly, while BLEU-2 and ROUGE do not. These examples are drawn randomly (i.e.\@ no cherry-picking) from the examples where {\sc adem} outperforms BLEU-2 and ROUGE (as defined in the text). {\sc adem} is able to correctly assign high scores to short responses that have no word-overlap with the reference response. The bars around $|$metric$|$ indicate that the metric scores have been normalized. }
\vspace{-4mm}
\end{table*}

Finally, we show the behaviour of {\sc adem} when there is a discrepancy between the lengths of the reference and model responses. In \citep{liu2016not}, the authors show that word-overlap metrics such as BLEU-1, BLEU-2, and METEOR exhibit a bias in this scenario: they tend to assign higher scores to responses that are closer in length to the reference response.\footnote{Note that, for our dataset, BLEU-2 almost exclusively assigns scores near 0 for both $\Delta w \leq 6$ and $\Delta w > 6$, resulting in a p-value \>0.05.} However, humans do not exhibit this bias; in other words, the quality of a response as judged by a human is roughly independent of its length. In Table \ref{tab:delw}, we show that {\sc adem} also does not exhibit this bias towards similar-length responses. 

\end{document}